\def\year{2021}\relax
\documentclass[letterpaper]{article} 
\usepackage{aaai21}  
\usepackage{times}  
\usepackage{helvet} 
\usepackage{courier}  
\usepackage[hyphens]{url}  
\usepackage{graphicx} 
\usepackage{natbib}  
\usepackage{caption} 
\usepackage{amsfonts} 
\usepackage{bm} 
\usepackage{amsmath} 
\usepackage{dsfont}

\graphicspath{ {./images/} }

\title{StatEcoNet: Statistical Ecology Neural Networks \\ for Species Distribution Modeling}
\author{
    Eugene Seo, \textsuperscript{\rm 1}
    Rebecca A. Hutchinson, \textsuperscript{\rm 1,2}
    Xiao Fu, \textsuperscript{\rm 1}
    Chelsea Li, \textsuperscript{\rm 1} \\ 
    Tyler A. Hallman, \textsuperscript{\rm 4}
    John Kilbride, \textsuperscript{\rm 3}
    W. Douglas Robinson \textsuperscript{\rm 2}
    \\
}
\affiliations{
    \textsuperscript{\rm 1}School of Electrical Engineering and Computer Science,
    \textsuperscript{\rm 2}Department of Fisheries and Wildlife,\\
    \textsuperscript{\rm 3}College of Earth, Ocean, and Atmospheric Sciences\\
    Oregon State University, Corvallis, OR 97331\\
    
    \textsuperscript{\rm 4}Monitoring Department, 
    Swiss Ornithological Institute, Sempach, Switzerland\\
    
    \{seoe,rah,xiao.fu,lichel\}@oregonstate.edu, tyler.hallman@vogelwarte.ch, \{kilbridj,douglas.robinson\}@oregonstate.edu
}

\begin{document}

\maketitle

\begin{abstract}
This paper focuses on a core task in computational sustainability and statistical ecology: species distribution modeling (SDM). In SDM, the occurrence pattern of a species on a landscape is predicted by environmental features based on observations at a set of locations. At first, SDM may appear to be a binary classification problem, and one might be inclined to employ classic tools (e.g., logistic regression, support vector machines, neural networks) to tackle it. However, wildlife surveys introduce structured noise (especially under-counting) in the species observations. If unaccounted for, these observation errors systematically bias SDMs. To address the unique challenges of SDM, this paper proposes a framework called \texttt{StatEcoNet}. Specifically, this work employs a graphical generative model in statistical ecology to serve as the skeleton of the proposed computational framework and carefully integrates neural networks under the framework. The advantages of \texttt{StatEcoNet} over related approaches are demonstrated on simulated datasets as well as bird species data.
Since SDMs are critical tools for ecological science and natural resource management, \texttt{StatEcoNet} may offer boosted computational and analytical powers to a wide range of applications that have significant social impacts, e.g., the study and conservation of threatened species.
\end{abstract}

\section{Introduction}
Estimating species distributions across a landscape is a fundamental problem in ecology. Species distribution models (SDMs) learn the relationship between the species of interest and a set of environmental features (e.g., elevation, land cover) from data collected at points on the landscape \cite{Elith2009,Franklin2010}. The species data may come from historical records~\cite{Elith2006}, professional surveys~\cite{Betts2008}, or volunteers in community science projects \footnote{Also referred to as ``citizen science" projects/data.}~\cite{Fink2010}. The environmental data may be collected \textit{in situ} or linked to the observation points post hoc (e.g., via remote sensing~\cite{Shirley2013}). SDMs are critical tools for both scientific inquiry and natural resource management, as they are employed to investigate how environmental features define species habitat and predict where species can persist successfully~\cite{Araujo2012}. 

At first glance, species distribution modeling may appear to be a straightforward machine learning (ML) problem, but the complex nature of ecological systems and the noise-prone data acquisition process entail unique challenges that are not addressed in conventional ML frameworks. First, data on species distributions are persistently plagued by \textit{imperfect detection}, in which some individuals of the species are missing from the data because of poor observation conditions, species behavioral traits, and/or limited survey efforts. 
Second, species respond to their environment in complex ways, so models of this process must handle many input variables and represent nonlinear relationships.
Third, models must be as interpretable as possible in order to translate their conclusions to meaningful scientific insights and effective management policies. 
Finally, SDMs are often built from smaller datasets than some other ML domains, with hundreds rather than thousands or millions of examples.

Classic approaches like regression models fail to capture systematic imperfect detection~\cite{Guillera-Arroita2014,Lahoz-monfort2014}. Instead, a family of latent variable models has been developed in statistical ecology to account for error in the observation process~\cite{Royle08,MacKenzie18}. This family originated with \textit{occupancy models}, in which the species \textit{occupancy} (occurrence) at a set of sites is represented with binary latent variables, and the species observations depend on occupancy status as well as a \textit{detection probability}~\cite{mackenzie2002estimating}.
In these models, the latent variables are of great scientific interest. Understanding how the environment determines occupancy may not only advance ecological research, but also assist policy decisions---e.g., making conservation policies for threatened species. 
Various extensions to this latent variable modeling framework have been introduced (e.g., 
with count-valued latent variables~\cite{royle2004n}), but this paper focuses on the occupancy model as a representative example. These models are often used within a classic statistical paradigm, where the probabilities of occupancy and detection are linked to features with regression functions, and models are selected with criteria like AIC.
This framework provides an effective approach to imperfect detection. However, it has limited modeling capacity due to the use of the linear regression model and thus struggles to model complex (i.e., highly nonlinear) relationships in high-dimensional feature spaces.

To handle the challenge of complexity in species' environmental responses, many ecologists have turned to machine learning~\cite{Elith2006}. In particular, boosted regression trees (BRT) and random forests (RF) are popular for their flexibility and predictive power~\cite{elith2008working,Cutler2015}; neural networks (NN) are an obvious alternative but have been explored less in this domain. Tree-based methods incorporate mechanisms for interpreting the model, such as feature importance metrics and partial dependence plots. However, these models treat species distribution modeling as a standard supervised classification problem without regard to the effects of imperfect detection; ignoring imperfect detection can cause systematic underestimation of species distributions. Furthermore, the effects of the features cannot be clearly separated into occupancy and detection components.

\subsubsection{Contributions.} 
This work puts forth a statistical ecology-inspired neural network model to address the above challenges. Our specific contributions are as follows. First, we propose a {\it statistical ecology-based neural network model} (\texttt{StatEcoNet}). The framework combines the statistical occupancy modeling approach that captures imperfect detection with neural networks that capture nonlinear relationships between the environment and species. We also introduce an easy-to-implement regularization strategy for selecting relevant features for the occupancy and detection sub-models, instead of requiring the user to specify these assignments. Specifically, we propose to use a group-sparsity regularization in the first layers of the NNs in \texttt{StatEcoNet}, thereby clearly indicating importance of the features to the two sub-models of the occupancy framework. Note that group-sparse predictors are often considered in linear regression and compressive sensing \cite{jenatton2011structured}, but have not been considered in interpretable ecological system neural modeling. We show advantages of \texttt{StatEcoNet} over alternative approaches on simulated data as well as a case study modeling five bird species.

\subsubsection{Prior Work.} 
Two pieces of prior work have attempted to address combinations of these challenges. First, nonlinear models have been incorporated into occupancy models using boosted regression trees (called \texttt{OD-BRT})~\cite{hutchinson2011incorporating}. That approach jointly fits two tree ensembles which are linked through an objective function that corresponds to the occupancy model likelihood. This addresses imperfect detection while automatically representing complex relationships to the features, but our experiments with this method indicate that it is difficult to tune properly and that it does not scale well to large datasets. Other recent work has also found that algorithms for learning BRT models are computationally intensive and can experience numerical instability~\cite{NIPS2017_6907}. Second, recent work incorporates nonlinear models into occupancy models with neural networks instead of BRTs~\cite{joseph2020neural}. However, it combines the features into a single network to model occupancy and detection, which limits interpretability.

\section{Problem Statement}
Consider a typical SDM setting where we are given binary observations (i.e., species detection or non-detection) made by observers at different sites. More formally, we define the following notation. The $t$th (where $t\in[T]$) observation at site $i$ (where $i\in[M]$) is denoted by $y_{it}$. Note that $y_{it}\in\{0,1\}$, where $y_{it}=1$ means that the target species was observed at site $i$ in the $t$th observation made, and $y_{it}=0$ otherwise. For every observation, survey-specific features (e.g., temperature, time of day of the observation) are recorded and collected in ${\bf w}_{it}\in\mathbb{R}^K$. Every site is characterized by a number of site-specific features (e.g., elevation, forest type), which are collected in ${\bf x}_i\in\mathbb{R}^{J}$. The objective is to determine the occurrence pattern of the species from the observations and the site and survey features. After the relationship is learned, the model can be used to predict species observations for new sites. In many studies, it is also critical to interpret how site features affect the species---i.e., to identify the environmental drivers of its distribution.

\subsubsection{Conventional Machine Learning Solution.} 
From an ML viewpoint, it is tempting to treat the $y_{it}$ as binary labels and concatenate the features to form ${\bf  u}_{it}=[{\bf w}_{it}^\top,{\bf x}_i^\top]^\top \in\mathbb{R}^{J+K}$. Then, an {\it empirical risk minimization} (ERM)-type formulation could be employed:
\begin{equation}\label{eq:ML}
    \min_{\bm \theta}~\sum_{i=1}^M\sum_{t=1}^T{\cal L}(y_{it}||f_{\bm \theta}({\bf u}_{it})),
\end{equation}
where $f_{\bm \theta}(\cdot):\mathbb{R}^{K+J}\rightarrow \mathbb{R}$ is any established model in ML (e.g., logistic regression, neural networks), $\bm \theta$ collects the model parameters (e.g., neural network weights), and ${\cal L}(x||y)$ is a loss function
(e.g., least squares, cross entropy).

\subsubsection{Challenges.} 
The ML solutions summarized in \eqref{eq:ML} seem reasonable, but the unique challenges of SDM may hinder performance. First, imperfect detection implies that some reports do not reflect the true status of the species at the site (e.g., when they are silent, hiding, or camouflaged), so these data contain structured noise. The probability of detecting a species varies across sites and surveys and is affected by numerous factors when conducting field surveys. Second, unlike classic applications of ML to `big data,' many ecological datasets are collected under substantial resource constraints. It is common to analyze hundreds of sites, in contrast to millions of images. Hence, exclusively data-driven ML models, e.g., deep neural networks, may not be applicable. 

To summarize, a completely data-driven complex ML model like deep neural networks may not be a viable solution for SDM. Nonetheless, neural networks offer appealing learning capacity in the presence of complex nonlinear transformations in the data generation process---and their companion algorithms balance modeling complexity, computational efficiency/stability, and generalization performance. These nice properties should be capitalized upon in SDM (e.g., for modeling the complex relations between site features and species distributions as well as the survey features and species detectability) with special attention paid to the ubiquity of missed detections and data scarcity challenges---this is the starting point of our work.

\section{Proposed Framework}
To address the challenges of applying advanced neural network-based learning techniques in SDM, we propose to integrate neural network-based nonlinear modeling with classic graphical generative models in statistical ecology. In a nutshell, the statistical model captures the effect of imperfect detection. The neural networks overcome model discrepancies that are often over-simplified in classic ecology models. This way, the neural networks are only responsible for handling the most challenging parts in the statistical model, while leaving the `well-understood' part to the classic model based approach. This reduces the complexity of the network and makes the learning process more efficient. 

\subsection{Preliminaries: The Occupancy Model}
\begin{figure}[ht]
    \centering
    \includegraphics[width=.45\linewidth]{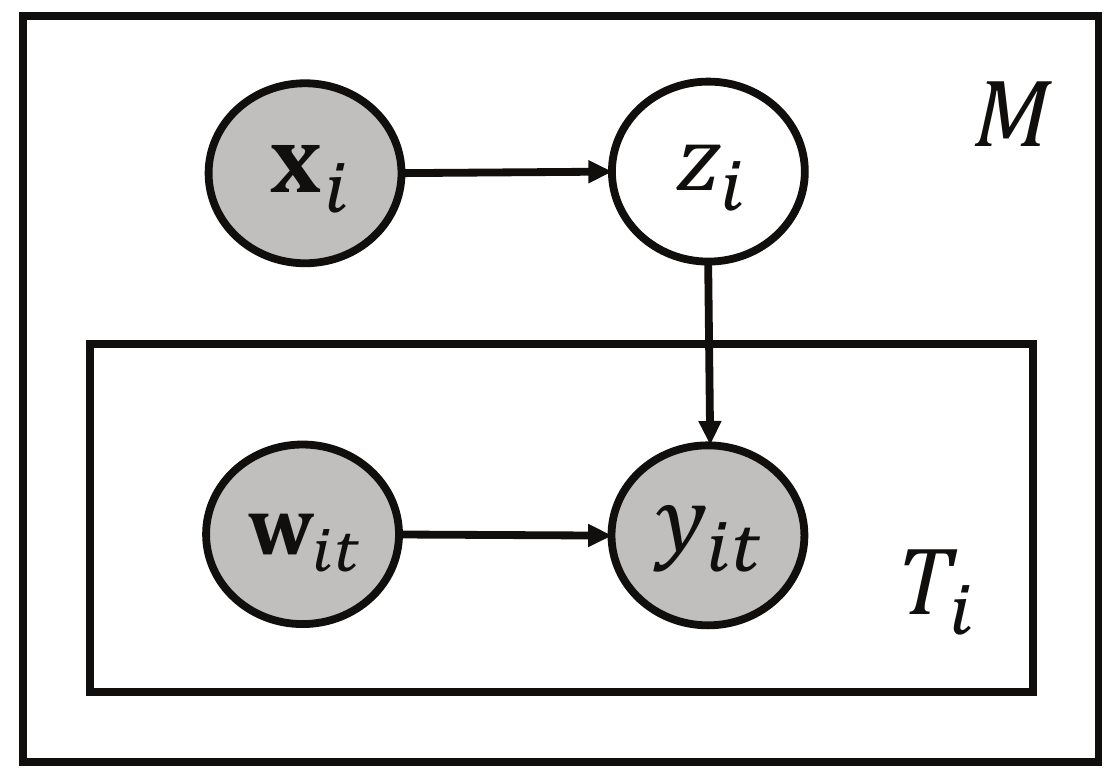}
    \caption{Graphical occupancy model. 
    $z_i \in \{0,1\}$ denotes latent species occupancy at site $i$ (of $M$ total) and $y_{it}\in \{0,1\}$ denotes the $t$th observation (of $T_i$ total). $\mathbf{x}_i$ and $\mathbf{w}_{it}$ are site and survey features, respectively.}
    \label{fig:latent_variable_model}
\end{figure}

The backbone of our proposed \texttt{StatEcoNet} is a widely accepted model in statistical ecology called the occupancy model~\cite{mackenzie2002estimating,MacKenzie18}. The graphical representation of the latent variable model is shown in Fig.~\ref{fig:latent_variable_model}. For each site $i = 1,...,M$, the biological model connects the true species occupancy status, $z_i \in \{0,1\}$ to site features $\mathbf{x}_{i}$ through an occupancy probability $o_i$. The key advance of the occupancy model over the approach of \eqref{eq:ML} is the introduction of the latent variable $z_i$ to capture the \textit{true occupancy status} of the species at site $i$. The acquired data $y_{it}$'s are treated as noisy observations of $z_i$, since they are influenced by imperfect detection. Letting each site contain $t = 1,...,T_i$ replicate surveys, the observation model links survey features $\mathbf{w}_{it}$ to a detection probability $p_{it}$. Note that introducing a detection probability that is associated with each observation is critical for SDM, since it explicitly models systematic under-counting. This model is intuitively and scientifically appealing, since it separates the causes for occupancy and detection; interpreting these effects separately is valuable in many ecological studies.

In the generative model, the observed data ${y}_{it}$ are produced by drawing from the occupancy Bernoulli and multiplying the result by the detection probability, i.e., \[ y_{it} \sim {\rm Bernoulli}(z_id_{it}). \] This encodes the assumption that unoccupied sites are always observed to be unoccupied, but that occupied sites might also be observed to be unoccupied. However, while it is clear that each observation $y_{it}$ is affected by both the true occupancy $z_i$ and the detection probability $d_{it}$, it is less clear how the site features ${\bf x}_i$ (resp. survey features ${\bf w}_{it}$) affect $z_i$ (resp. $d_{it}$). In classical applications of occupancy models, linear models map ${\bf x}_i$ and ${\bf w}_{it}$ to occupancy probability $o_i$ and detection probability $d_{it}$, respectively, through a linear logit modeling strategy~\cite{MacKenzie18}; i.e., 
\begin{align}\label{eq:linear}
    o_i &= \frac{\exp({\bf x}_i^\top\bm \alpha)}{1+\exp({\bf x}_i^\top\bm \alpha)},\quad
   d_{it} = \frac{\exp({\bf w}_{it}^\top\bm \beta)}{1+\exp({\bf w}_{it}^\top\bm \beta)},
\end{align}
where $\bm \alpha\in\mathbb{R}^K$ and $\bm \beta\in\mathbb{R}^{J}$ are model parameters to be estimated.
The true occupancy has probability $o_i$, i.e.,
\[     z_i \sim {\rm Bernoulli}(o_i). \]
This framework makes sense, but the linear models are over-simplified for complex ecological systems. 

\subsection{Integrating Neural Networks into the Framework}
In this work, we propose to use {\it two} neural networks to model the relations between ${\bf x}_{i}$ and $o_i$ as well as ${\bf w}_{it}$ and $d_{it}$ in \eqref{eq:linear}. Our motivation is not to replace the well established graphical model in Fig.~\ref{fig:latent_variable_model} by a completely data-driven neural network (as in \eqref{eq:ML}), but to leverage the power of neural networks to fill the `modeling gap' of the graphical model. For statistical ecologists, this is perhaps the most natural way of integrating neural networks into SDM.

\begin{figure}[ht]
    \centering
    \includegraphics[width=0.85\linewidth]{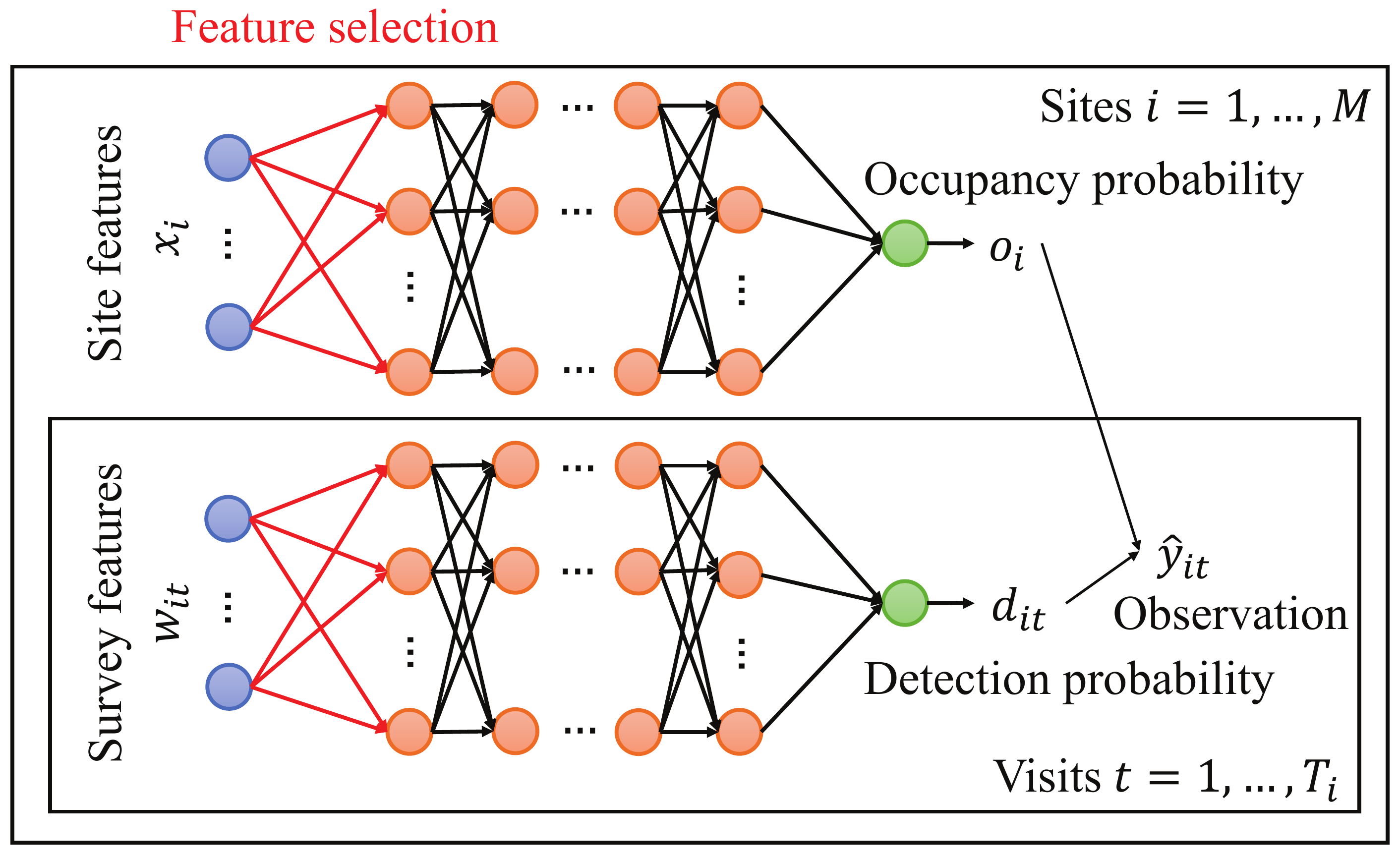}
    \caption{Proposed model (\texttt{StatEcoNet}) framework.}
    \label{fig:proposed_model}
\end{figure}

Specifically, we introduce two neural networks \[F(\cdot):\mathbb{R}^K\rightarrow \mathbb{R},~ G(\cdot):\mathbb{R}^{J}\rightarrow \mathbb{R}\] as shown in Fig.~\ref{fig:proposed_model}. The first neural network $F(\mathbf{x}_i)$ predicts the occupancy probability from the given site features $\mathbf{x}_i$. The second neural network predicts the detection probability from given survey features $\mathbf{w}_{it}$. 

We employ fully connected networks to express $F$ and $G$:
\begin{subequations}
\begin{align}
F({\bf x}_i;{\bm \theta}_F)&={\bf u}_L^\top \bm \sigma({\bf U}_{L-1}\bm \sigma(\cdots \bm \sigma({\bf U}_1{\bf x}_i) )) , \label{eq:F} \\
G({\bf w}_{it};\bm \theta_G)&={\bf v}_L^\top \bm \sigma({\bf V}_{L-1}\bm \sigma(\cdots \bm \sigma({\bf V}_1{\bf w}_{it}) )). \label{eq:G}
\end{align}
\end{subequations}
In \eqref{eq:F}, 
${\bf U}_{\ell}\in\mathbb{R}^{K_{\ell}\times K_{\ell-1}}$ 
is the network weight in the $\ell$th layer where $K_{\ell}$ is the number of neurons of the $\ell$th layer (and we define $K_0=K$). The output layer has a combining vector ${\bf u}_L\in\mathbb{R}^{K_{L-1}}$ 
that maps the output to a scalar. The weights in \eqref{eq:G} are defined in the same way. We also define $\bm \theta_F$ and $\bm \theta_G$ as the collections of the network parameters of $F$ and $G$, respectively. The function $\bm \sigma(\cdot)$ applies onto every element of its input individually. We employ the popular rectified linear unit (ReLU) function as our activation function. With the neural networks defined, the occupancy and detection probabilities can be re-expressed as follows:
\begin{subequations}
\begin{align}
 & ~o_i = \frac{\exp(F({\bf x}_i;\bm \theta_F))}{1+\exp(F({\bf x}_i;\bm \theta_F))}  \label{eq:occupancy},\\
 & ~d_{it} =\frac{\exp(G({\bf w}_{it};\bm \theta_G))}{1+\exp(G({\bf w}_{it};\bm \theta_G))}. \label{eq:detection}
\end{align}
\end{subequations}
With the above construction and the overall graphical model, we define a maximum likelihood estimation problem whose log-likelihood function can be expressed as:
\begin{equation}
    \begin{aligned}
    &\log{\cal L}= \sum_{i=1}^M \log{\cal L}_i \\
    =&\sum_{i=1}^M \log\left(  o_i \prod_{t=1}^{T_i} \left[d_{it}^{y_{it}} (1-d_{it})^{1-y_{it}}\right] + (1-o_i)\kappa_i  \right),
    \end{aligned}
\end{equation}
where $\kappa_i$ is an indicator function defined as $\mathds{1} \left(\sum_{t=1}^{T_i} y_{it} = 0 \right)$, in which $\mathds{1}(\cdot)$ is 1 if the observations at a site were all zero and 0 otherwise.
In the above, we have followed the derivation of \cite{hutchinson2011incorporating} to reach the expression of ${\cal L}_i$ from 
${\cal L}_i 
=\sum_{z\in\{0,1\}} {\sf Pr}(z_i = z)\prod_{t=1}^{T_i}{\sf Pr}(y_{it} \vert z_i =z)
=\sum_{z\in\{0,1\}} o_i^z(1-o_i)^{1-z}\prod_{t=1}^{T_i} (zd_{it})^{y_{it}}(1-zd_{it})^{1-y_{it}}$.

\subsection{Feature Selection via the $\ell_{2,1}$-Norm}
On top of this structure, we incorporate regularization terms into our model in order to identify features that significantly impact each of the model probabilities. There has been little work on incorporating the feature selection process into neural network models. Instead, most prior work selects relevant features as a preprocessing before learning the neural network model~\cite{cheng2020spatio}. Here, we add the $\ell_{2,1}$-norm into our \texttt{StatEcoNet} to reveal which features impact the occupancy and detection probabilities.

The mixed $\ell_{2,1}$-norm (also denoted as $\ell_2/\ell_1$-norm) is a matrix norm used for robust optimization problems that promotes sparsity of the matrix columns. It thus has widely been used in signal and image processing to handle noise and outliers~\cite{steffens2018compact}. Accordingly, the $\ell_{2,1}$-norm has been considered an approach for feature selection~\cite{jenatton2011structured}. The $\ell_{2,1}$-norm of 
${\bf U}_{\ell}\in\mathbb{R}^{K_{\ell}\times K_{\ell-1}}$
is defined as
\begin{align}
\Vert {\bf U}_{\ell} \Vert_{2,1} &= \sum_{j=1}^{K_{\ell-1}} ( \sum_{i=1}^{K_{\ell}} \vert u_{ij} \vert^{2})^{1/2} = \sum_{j=1}^{K_{\ell-1}} \Vert {\bf U}_{\ell}(:,j) \Vert_2.
\end{align}
The $\ell_{2,1}$-norm behaves like an $\ell_{1}$-norm on a vector for providing a sparse solution to the columns of a matrix. That is, the parameter matrix is regularized with the $\ell_{2,1}$-norm minimization in order to discover important features. We introduce this mixed $\ell_{2,1}$-norm into the first input layer of both neural networks, where the parameter matrix is connected to the input features as shown in Fig.~\ref{fig:proposed_model}.

Our regularized loss function is given by
\begin{align}
-\sum_{i=1}^{M} \log{\cal L}_i + \lambda_F \Vert {\bf U}_1 \Vert_{2,1} + \lambda_G \Vert {\bf V}_1 \Vert_{2,1},
\label{eq:regLoss}
\end{align}
where $\lambda_F$ and $\lambda_G$ are regularization weights for the occupancy and detection features, respectively. Thus, the goal of the learning algorithm is to minimize the negative log-likelihood of our occupancy model as well as the $\ell_{2,1}$-norms.
 
\subsection{Training via Subgradient}
A benefit of using neural network based modeling is that the computational tools for neural network-related optimization problems are well-developed. In particular, using a subgradient-based framework and leveraging effective backpropagation-based subgradient computation, the per-iteration complexity of the algorithm is appealing. The maximum likelihood estimation problem is unconstrained, and thus a simple subgradient descent algorithm can be naturally employed. Since the three terms in \eqref{eq:regLoss} are all non-differentiable (since the neural networks use the ReLU activation function), subgradient should be used, instead of gradient. More algorithmic details are in the supplement.

\section{Experiment Design}
We evaluated our model with both simulated and avian point count data. 
We compared our models with three other approaches, each tuned individually for a peak-to-peak comparison. The code and supplementary material are available at https://github.com/Hutchinson-Lab/StatEcoNet-AAAI21.

\subsection{Synthetic Data}
We simulated data to evaluate the models' ability to predict probabilities and observations as well as discover important features under the assumed model. We constructed ten features each for the occupancy and detection components, but only the first five features had non-zero coefficients (i.e., each sub-model had five irrelevant features). This setting is for testing the effectiveness of the feature selection layer in \texttt{StatEcoNet}. We generated data with both linear and nonlinear effects of the features on the occupancy and detection probabilities. In total, we simulated training and validation sets from the eight combinations of $M \in \{100, 1000\}$, $T \in \{3, 10\}$, and feature-occupancy/detection model $\in\{\text{linear,~nonlinear}\}$. Testing sets always had $M=1000$ for more robust performance estimates. More detailed simulation settings can be seen in the supplemental material.

\subsection{Avian Point Count Data}
We also analyzed data on bird distributions to evaluate the proposed method on real-world datasets. We analyzed 10,845 5-minute point count bird surveys extracted from the Oregon 2020 dataset collected in Oregon, United States \cite{oregon2020}. Surveys were conducted during the bird breeding season (May 15-July 10) by trained field ornithologists from 2011 to 2019. We selected five common Oregon species for this analysis. Common Yellowthroat (\textit{Geothlypis trichas}), Eurasian Collared-Dove (\textit{Streptopelia decaocto}), Pacific Wren (\textit{Troglodytes pacificus}), Song Sparrow (\textit{Melospiza melodia}), and Western Meadowlark (\textit{Sturnella neglecta}) vocalize frequently during the breeding season and have conspicuous, easily identifiable vocalizations. These species have very different habitat preferences (see supplement for more details). Tab.~\ref{tab:species_stat_percentage} shows statistics of our datasets (i.e., the percentage of the sites and surveys with positive observations of the species).
\begin{table}[ht]
    \centering
    \begin{tabular}{ |c|c|c| } 
        \hline
        Species & \multicolumn{2}{c|}{Percent observed} \\
         & sites & surveys \\ 
        \hline
        \hline
        Common Yellowthroat (COYE)   & 19.5\% &  10.7\% \\ 
        Eurasian Collared-Dove (EUCD) & 14.0\% &  8.2\% \\ 
        Pacific Wren (PAWR)          & 24.3\% &  14.5\% \\ 
        Song Sparrow (SOSP)          & 45.8\% &  27.8\% \\ 
        Western Meadowlark (WEME)    & 15.5\% &  12.2\% \\ 
        \hline
    \end{tabular}
    \caption{Species analyzed and the percent of sites (of 942 total) and surveys (of 942 sites $\times$ 3 visits per site $=$ 2,826 total) with positive observations of the species.}
    \label{tab:species_stat_percentage}
\end{table}

\begin{figure}[ht]
    \centering
    \includegraphics[width=0.9\linewidth]{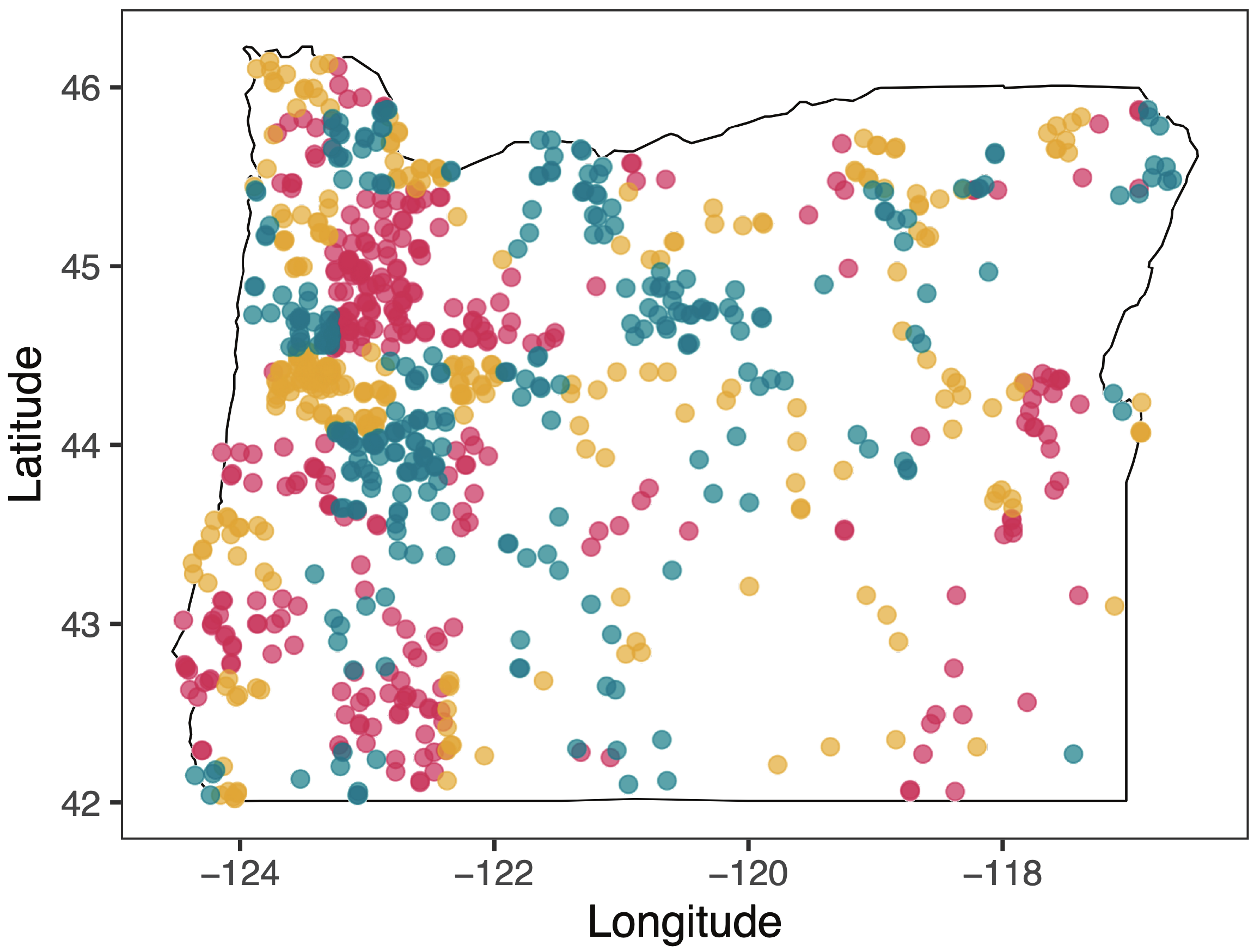}
    \caption{Map of the survey sites over Oregon, United States. Each site had at least three surveys. Colors indicate assignment of sites into three folds (training, validation, test) for Western Meadowlark.}
    \label{fig:real_sites_folds}
\end{figure}

Before fitting models, we acquired site and survey features, grouped observations into sites, and divided the data for cross-validation. We constructed 28 environmental features describing the sites from Landsat satellite image composites (details in supplement). The observation-related features were year, day, and time of observation, to capture time-varying detectability. For bird datasets, we consider both environmental and observation-related features as detection features because the site-specific information can affect species detectability. Though Oregon 2020 did not explicitly pre-define sites with multiple visits, its clustered sampling design simplified survey-to-site-assignment. We pre-processed the data by excluding sites that were only surveyed one or two times, and for sites visited more than three times, we randomly selected three surveys. This resulted in a total of 942 sites. We divided these data into three spatially distinct cross-validation folds \cite{Valavi2018}. The site distribution and fold assignments are shown in Fig.~\ref{fig:real_sites_folds}. 

\subsection{Performance Metrics}
We evaluated model quality along several dimensions. We measured the Pearson correlation coefficient between the true and estimated model probabilities for the simulated datasets. For predicting held-out observations, we measured both the area under the Receiver Operating Characteristic Curve (AUROC) and the area under the Precision-Recall Curve (AUPRC). Note that the probability of a positive observation is the product of the occupancy and detection probabilities. In this case, AUPRC may be preferred over AUROC since AUPRC is better suited to class-imbalanced data~\cite{davis2006relationship} (Tab.~\ref{tab:species_stat_percentage}). With synthetic datasets, we compared the features that each model selects based on its own relative influence scores against the truly relevant features in the data-generation procedure. When applying \texttt{StatEcoNet} to the avian datasets, we present the $\ell_{2}$-norms of ${\bf U}_1(:,j)$ and ${\bf V}_1(:,k)$ as the indicators of the importance of the features $[{\bf x}_i]_j$ and $[{\bf w}_{it}]_k$, respectively. In \texttt{OD-BRT}, we present the number of times that each feature was selected as a split variable as the indicator of feature importance. Finally, we compare the performance of the models by measuring  training time. We repeated experiments 5 times for synthetic datasets and 10 times for avian point count datasets, and summarized the performance evaluation metrics with mean and standard deviation values.
\begin{table*}[t]
    \centering
    \begin{tabular}{ |c|c|c|c|c|c| } 
        \hline
        Method & Training Time & Occ.Prob.Corr. & Det.Prob.Corr. & AUPRC & AUROC \\ 
        \hline
        \hline
        \texttt{OD-LR}      &  \textbf{3.66} $\pm$ 3.11 s            & 0.05 $\pm$ 0.001          & 0.01 $\pm$ 0.001          & 0.32 $\pm$ 0.002          & 0.51 $\pm$ 0.001\\
        \texttt{OD-1NN}     & 30.3 $\pm$ 5.15 s    & \textbf{0.84} $\pm$ 0.01  & 0.004 $\pm$ 0.003         & 0.39 $\pm$ 0.004          & 0.61 $\pm$ 0.01 \\
        \texttt{OD-BRT}     & 320 $\pm$ 60.6 s              & 0.83 $\pm$ 0.01           & \textbf{0.97} $\pm$ 0.002 & \textbf{0.53} $\pm$ 0.003 & 0.72 $\pm$ 0.002\\
        \texttt{StatEcoNet} & 94.2 $\pm$ 17.5 s             & \textbf{0.84} $\pm$ 0.01  & \textbf{0.97} $\pm$ 0.003 & \textbf{0.53} $\pm$ 0.001 & \textbf{0.73} $\pm$ 0.003\\
        \hline
    \end{tabular}
    \caption{Performance metrics (mean $\pm$ st. dev.) on simulated data with $M=1000$, $T=10$, and nonlinear relationships. }
    \label{tab:syn_data_res}
\end{table*}

\begin{table*}[t]
    \centering
    \begin{tabular}{ |c|c|c|c|c|c|c|c|c|c|c| } 
        \hline
         & \multicolumn{2}{c|}{COYE} & \multicolumn{2}{c|}{EUCD} & \multicolumn{2}{c|}{PAWR} & \multicolumn{2}{c|}{SOSP} & \multicolumn{2}{c|}{WEME} \\ 
        Method & mean & st.dev  & mean & st.dev  & mean & st.dev  & mean & st.dev  & mean & st.dev \\
        \hline
        \hline
        \texttt{OD-LR}      & 0.375 & 0.0614 & 0.208 & 0.0462 & 0.474 & 0.0382 & 0.563 & 0.0230 & 0.559 & 0.1320 \\
        \texttt{OD-1NN}     & 0.376 & 0.0495 & 0.272 & 0.0462 & 0.461 & 0.0311 & 0.567 & 0.0311 & 0.545 & 0.1269\\
        \texttt{OD-BRT}     & 0.369 & 0.0458 & 0.183 & 0.0453 & 0.473 & 0.0348 & 0.558 & 0.0322 & \textbf{0.634} & 0.0665\\
        \texttt{StatEcoNet} & \textbf{0.383} & 0.0519 & \textbf{0.283} & 0.0610 & \textbf{0.496} & 0.0314 & \textbf{0.571} & 0.0210 & 0.593 & 0.1049\\
        \hline
    \end{tabular}
    \caption{AUPRC for the five species on predicting held-out observations. This quantity is what we can measure on these data, since we do not have ground truth for occupancy, but it is not as scientifically interesting. Performance differences are minor. }
    \label{tab:OR2020_data_res}
\end{table*}

\subsection{Baselines and Parameter Tuning}
We compared the proposed method to three baselines: \texttt{OD-LR}~\cite{mackenzie2002estimating}, 
\texttt{OD-BRT}~\cite{hutchinson2011incorporating}, and \texttt{OD-1NN}~\cite{joseph2020neural}. Note that the superior performance of latent variable models compared to standard ML methods (e.g., logistic regression and ensembles of trees) has been demonstrated in the prior work ~\cite{hutchinson2011incorporating}. In addition, the latent variable structure of the occupancy model is critical for scientific inference in ecology. Hence, we focus on comparing our proposed method against two non-NN latent variable models (\texttt{OD-LR} for linear model and \texttt{OD-BRT} for tree-based nonlinear model) and one alternative of NN-based latent variable model (\texttt{OD-1NN}). The parameter tuning strategies for the methods under comparison are as follows. For \texttt{StatEcoNet}, we selected the key parameters, i.e., learning rate, batch size, number of neurons per layer, and number of layers, from $\{0.01,0.001,0.0001\}$, $\{32,all\}$, $\{8,16,32,64\}$, and $\{1,3\}$, respectively, to maximize the AUPRC performance on the validation set. Similarly, we tuned all parameters for the baselines. For \texttt{OD-BRT}, we used Bayesian optimization~\cite{Snoek2012,rBayesOptPkg} to tune the shrinkage, bag fraction, tree depth, and number of trees since this method was computationally intensive. The input features of bird species data were normalized for all methods except \texttt{OD-BRT}, as trees based methods do not require this procedure. More details are in the  supplemental material.

\section{Results}
\subsection{Simulation Study}
Overall, \texttt{StatEcoNet} was more effective than the baseline methods on simulated data. The estimated occupancy and detection probabilities from \texttt{StatEcoNet} were more correlated with the true probabilities than estimates from the other methods. Tab.~\ref{tab:syn_data_res} shows results for a case where the relationships between features and the occupancy/detection probability are nonlinear, $M=1000$, and $T=10$; results for a variety of other settings are in the supplemental material. \texttt{OD-LR}'s performance suffers since it does not fit nonlinear relationships. \texttt{OD-1NN} estimated detection probabilities poorly, since the occupancy and detection sub-models were confounded in the single network, which may have made the network size unnecessarily large and the model hard to learn. \texttt{OD-BRT} estimated the target probabilities well on nonlinear data, but its training time was more than three times of that used for \texttt{StatEcoNet}. In addition, a perhaps unexpected observation is that \texttt{OD-BRT} struggled to learn the models when the feature-occupancy/detection probability models were linear (see details in the supplemental material). This may reflect difficulties with approximating lines by a `staircase' of axis-parallel splits.

\begin{figure}[ht]
    \centering
    \includegraphics[width=0.9\linewidth]{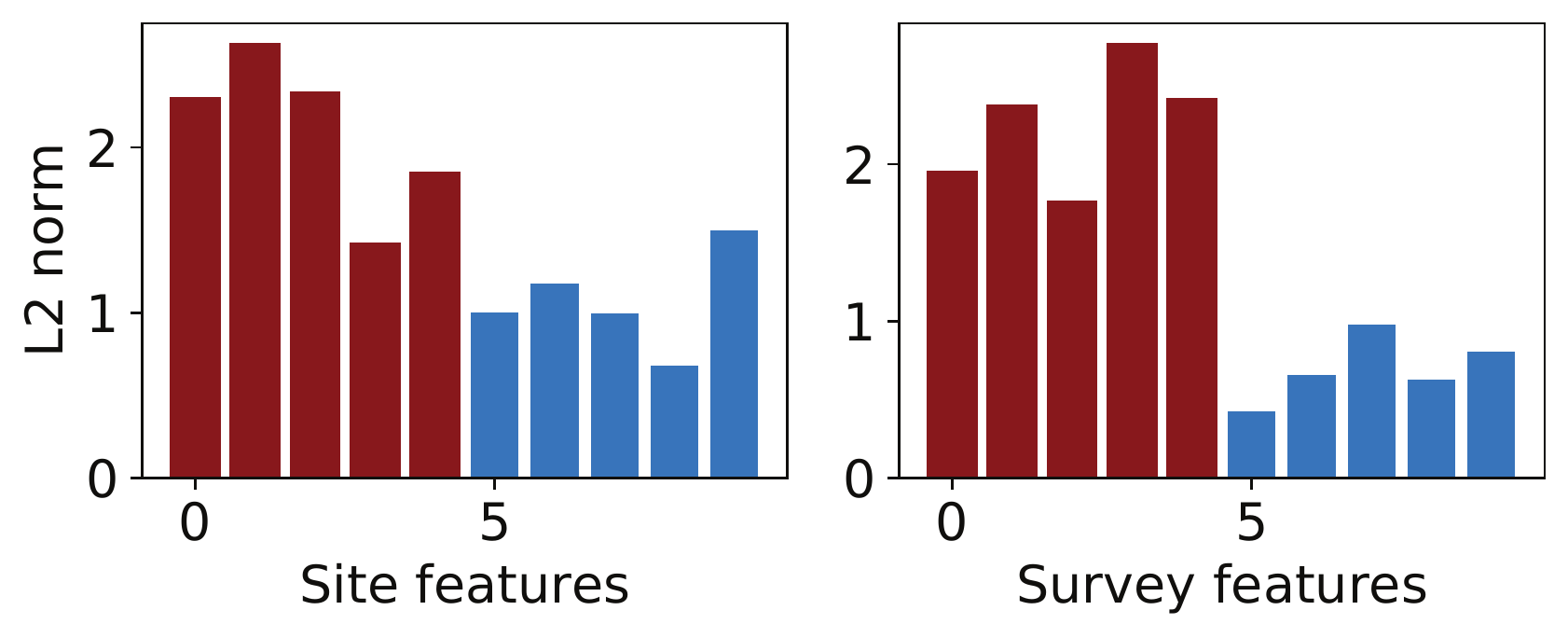}
    \caption{
    Selected features by $\texttt{StatEcoNet}$ for the synthetic dataset with $M$=1000, $T$=10, and nonlinear relationships. The dark red bars correspond to relevant features, and the blue bars irrelevant features. 
    }
\label{fig:syn_feature_all_methods}
\end{figure}

Fig.~\ref{fig:syn_feature_all_methods} shows the parameters learned by \texttt{StatEcoNet}: $\|{\bf U}_1(:,j)\|_2$ and $\|{\bf V}_1(:,k)\|_2$. \texttt{StatEcoNet} successfully identified most of the truly relevant features, as evidenced by the larger norms of the ${\bf U}_1(:,j)$ and ${\bf V}_1(:,k)$ corresponding to the relevant features (see more in supplement). This indicates the efficacy of the $\ell_{2,1}$-norm based regularization.

\subsection{Avian Point Count Study}
Performance evaluation in this study is challenging because ground truth for the model probabilities and feature importances are unknown. We can compare the methods' abilities to predict held-out observations ($y_{it}$), but it is important to note that \textit{occupancy}, not \textit{observation}, is of primary scientific interest in the model---precisely what we cannot evaluate directly. \texttt{StatEcoNet} outperforms the baseline methods on four of the five species tested (Tab.~\ref{tab:OR2020_data_res}).
\begin{figure*}
\centering
  \includegraphics[width=0.83\textwidth]{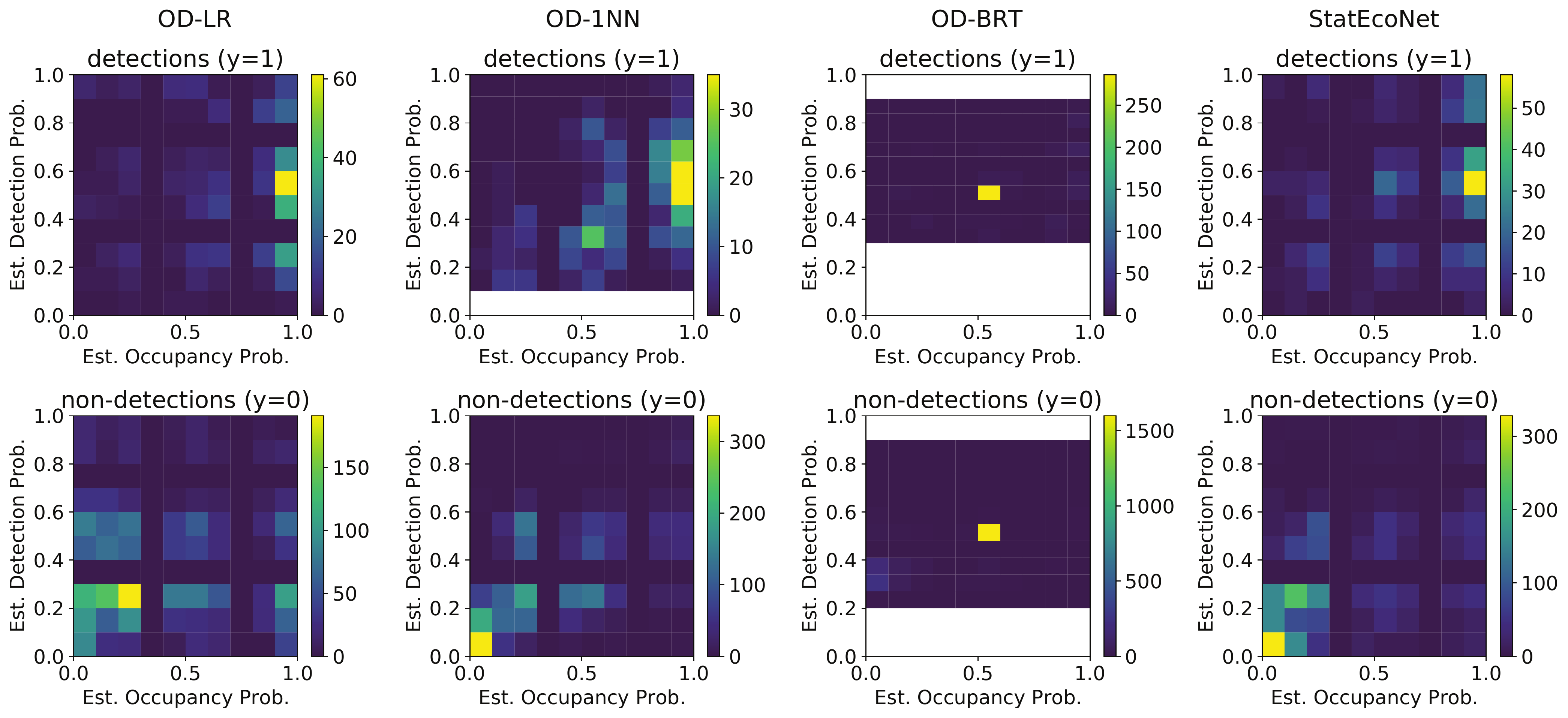}
  \caption{Histograms for Pacific Wren. 
  \texttt{OD-BRT} shows excessive clumping near 0.5. Ground truth is unknown, but \texttt{StatEcoNet} shows more realistic variability than \texttt{OD-LR} and \texttt{OD-1NN}.
  }
  \label{fig:PAWR_densities}
\end{figure*}

While impossible to validate exactly, it is illustrative to examine the occupancy and detection probabilities estimated by the different methods on these data. Recall that the predictions are a product of these probabilities (i.e., $\widehat{y}_{it} = \widehat{o}_i \widehat{d}_{it}$). Intuitively, if $\widehat{o}_i$ and $\widehat{d}_{it}$ are estimated correctly, the product $\widehat{o}_i \widehat{d}_{it}$ should be close to the observed events $y_{it}=1$ (detection) and $y_{it}=0$ (non-detection) on the test set. To use this intuition for evaluation, consider the Pacific Wren as an example. Fig.~\ref{fig:PAWR_densities} shows two-dimensional histograms of the learned occupancy probabilities $\widehat{o}_i$ and detection probabilities $\widehat{d}_{it}$ for each method, separated for the cases of positive and negative observations. The histogram is color coded, where brighter grids mean the corresponding events happen with higher frequencies. Ideally, a good model and learning algorithm would `light up' the upper right corner of the histogram for $y_{it}=1$ (first row in Fig.~\ref{fig:PAWR_densities}), which means that the estimated occupancy probability $\widehat{o}_i$ and detection probability $\widehat{d}_{it}$ can reproduce the held-out detected events. Similarly, for the $y_{it}=0$ events, an ideal method will make the bottom left corner `brighter' (second row in Fig.~\ref{fig:PAWR_densities}).

In Fig.~\ref{fig:PAWR_densities}, many of the \texttt{OD-BRT} model probability estimates are highly clustered around 0.5. This seems to indicate underfitting and is biologically unrealistic. The \texttt{OD-LR} and \texttt{OD-1NN} histograms did exhibit high frequencies at the upper right and lower left corners for the detection and non-detection events, respectively. However, the events and the learned models are concentrated in a relatively small number of grid cells, making the histograms spiky. 
This may be pathological since it models the observations with a small number of $\widehat{o}_{i}$ and $\widehat{d}_{it}$---but different sites and surveys may admit a large variety of $\widehat{o}_{i}$ and $\widehat{d}_{it}$ in reality. Hence, although these models could have good estimates for the product $\widehat{o}_i\widehat{d}_{it}$ (and thus similar AUPRCs to \texttt{StatEcoNet}), the individual estimates $\widehat{o}_{i}$ and $\widehat{d}_{it}$ may not be insightful for ecologists. Encouragingly, the histograms from \texttt{StatEcoNet} show more variability---the probabilities concentrate in the desired regions but also gracefully spread out. 

\begin{figure}[ht]
    \centering
    \includegraphics[width=\linewidth]{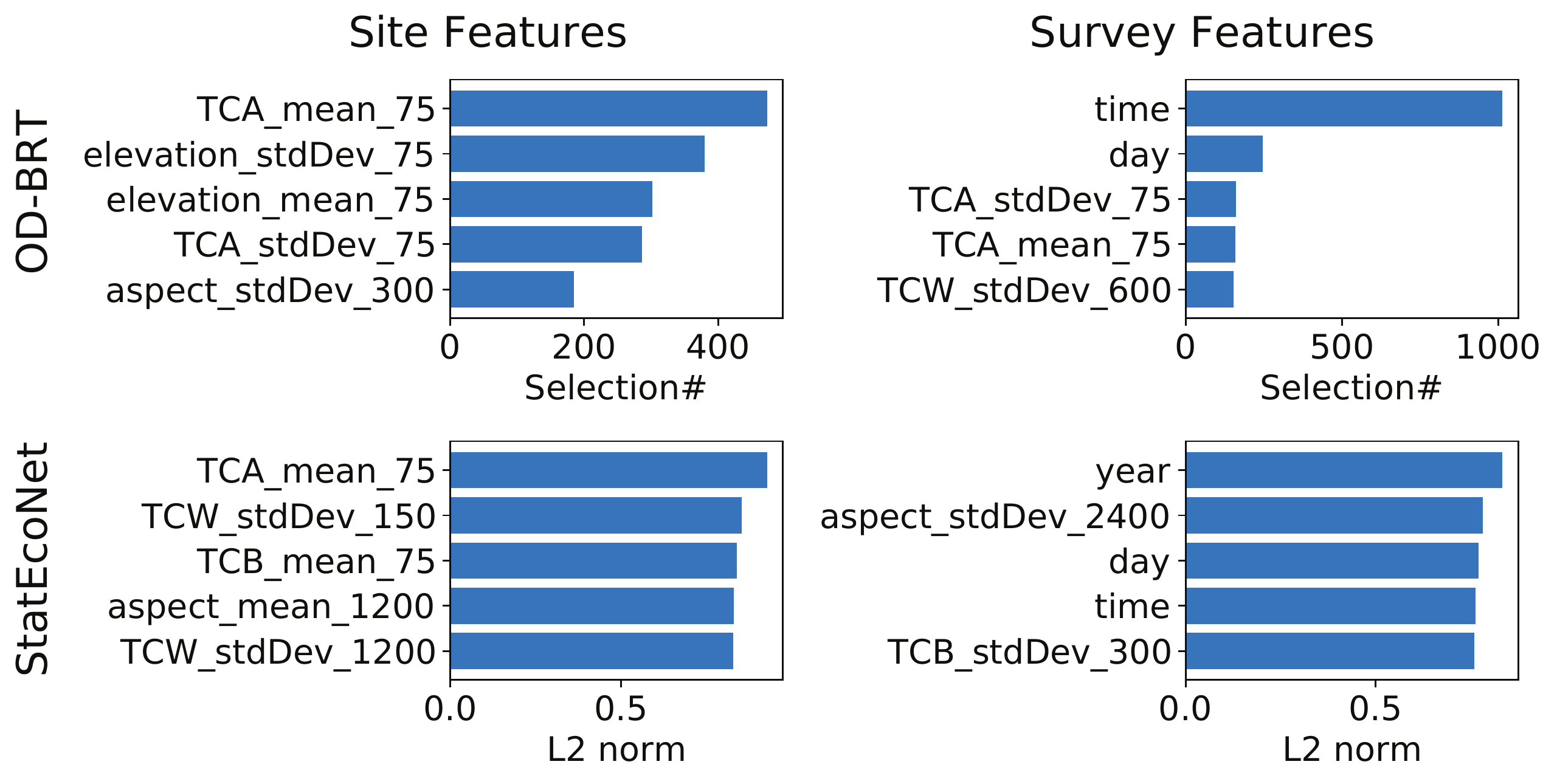}
    \caption{Comparison of feature importances from \texttt{OD-BRT} and \texttt{StatEcoNet} for Pacific Wren (fold 1). The plots on the left (right) show the important site (survey) features selected by each method.}
\label{fig:real_feature_all_methods}
\end{figure}

Finally, we examined feature importances on the bird datasets. Continuing with the Pacific Wren, Fig.~\ref{fig:real_feature_all_methods} shows the top five site and survey features selected by \texttt{OD-BRT} and \texttt{StatEcoNet}. Interestingly, \texttt{OD-BRT} emphasizes time almost exclusively in the detection model, while \texttt{StatEcoNet} blends the influence of the time-varying features with site-specific environmental features. For both methods, the most important feature was the mean of the land cover index, Tasseled Cap Angle (TCA) at the 75 meter scale. Since this species is found in wet forests with rich undergrowth on the forest floor, this feature may make intuitive sense because TCA is the land cover index that captures the information of both brightness and greenness of land cover, and thus it can indicate dense vegetation \cite{white2011history}. Even more promisingly, \texttt{StatEcoNet} selected another land cover index, Tasseled Cap Wetness (TCW) which represents wetness of area. The results for the other four species can be found in the supplemental material.

\section{Conclusion}
This paper contributes \texttt{StatEcoNet}, an interpretable computational framework to integrate the power of neural networks into statistical ecology models that account for the critical challenge of imperfect detection in species distribution modeling. Experiments on simulated datasets showed that \texttt{StatEcoNet} outperforms alternative approaches under various metrics for SDM. In particular, the examination of the learned probabilities and the selected features using real community science data on bird species exhibits intuitively pleasing and encouraging results. Since species distribution models are critical for science and conservation, and imperfect detection and model complexity are pervasive challenges for building these models, \texttt{StatEcoNet}'s ability to meet both of these challenges simultaneously has the potential for broad application and impact. In future work, we will aim to maximize this impact by analyzing more species datasets in collaboration with ecologists, improving the optimization procedure for sites with variable numbers of observations, and extending this framework beyond binary characterizations of species distributions.

\section{Acknowledgements}
We thank Laurel Hopkins, Jing Wang, and Mark Roth for helpful discussions and the anonymous reviewers for their valuable comments. This work was supported in part by the National Science Foundation under projects NSF IIS-1910118 and NSF ECCS 1808159.

\section{Ethics Statement}
This work has broad positive societal implications. 
Species distribution models are widely used to develop conservation and management policies for threatened species. In the midst of the sixth global mass extinction, effective actions for slowing biodiversity loss are critical for preserving our fellow inhabitants of Earth, as well as the ecosystem services they may provide to humans. The method contributed in this paper, \texttt{StatEcoNet}, offers new capacity in this area by simultaneously fitting nonlinear models for species occupancy and detection probabilities and identifying the features most important to each of those components. 
Our contributions capitalize on recent advances in neural networks to build more powerful SDMs. Our effort may lead to enhanced understanding of highly complex ecosystems and facilitate more effective conservation policies. This may be particularly critical in an age of drastic climate change, devastating hurricanes, and raging wildfire---the effects of which compound to threaten species persistence globally.

\bibliography{references.bib}
\end{document}


\maketitle

\clearpage

\tableofcontents

\clearpage

\pagebreak

\section{Introduction}
This document includes explanations and descriptions of our model training algorithm, generation of the synthetic data, parameter tuning process, and setup of our real-data experiments. In particular, the avian point count datasets of 5 bird species, including the full descriptions of the site and survey features, are detailed in this document. Additional simulation results and real-data experiments on four more bird species can also be found in this document. Sec.~\ref{sec:realdata} also presents discussions on the real-data experiments and insights revealed by the outputs of the algorithms from an ecological study viewpoint.

\section{Subgradient Algorithm}
Recall that the maximum likelihood estimation problem is as follows:
\begin{equation}
    \begin{aligned}
    \log{\cal L}= \sum_{i=1}^M \log{\cal L}_i 
    =\sum_{i=1}^M \log\left(  o_i \prod_{t=1}^{T_i} \left[d_{it}^{y_{it}} (1-d_{it})^{1-y_{it}}\right] + (1-o_i)\kappa_i  \right)
    \end{aligned}
\end{equation}
where $\kappa_i$ is a constant defined as $\kappa_i =  \mathbbm{1} \left(\sum_{t=1}^{T_i} y_{it} = 0 \right)$. The regularized version of our cost function is given by
\begin{align}
-\frac{1}{M}\sum_{i=1}^{M} \log{\cal L}_i + \lambda_F \Vert {\bf U}_1 \Vert_{2,1} + \lambda_G, \Vert {\bf V}_1 \Vert_{2,1}  
\label{eq:regLoss}
\end{align}
where the $\ell_2/\ell_1$ mixed norm for ${\bf Z}\in\mathbb{R}^{m\times n}$ is expressed as follows:
\[  \| {\bf Z} \|_{2,1}= \sum_{j=1}^n\| {\bf Z}(:,j)\|_2 .\]
As we mentioned, the mixed norm is often used in the literature for feature selection.
To put together, our optimization criteria can be summarized as
\begin{equation}\label{eq:ERM}
    \min_{\bm \theta_G,\bm \theta_F}~-\frac{1}{M}\sum_{i=1}^M \tilde{\cal L}_i(\bm \theta_G,\bm \theta_F) + \lambda_F\phi(\bm \theta_F) + \lambda_G\phi(\bm \theta_G),
\end{equation}
where
\begin{align*}
    \tilde{\cal L}_i(\bm \theta_G,\bm \theta_F)& = \log {\cal L}_i,\quad
    \phi(\bm \theta_F) = \|{\bf U}_1\|_{2,1},\quad
    \phi(\bm \theta_G)= \Vert {\bf V}_1 \Vert_{2,1}.
\end{align*}
The maximum likelihood estimation problem is unconstrained, and thus a simple subgradient descent algorithm can be naturally employed.
Since the three terms in \eqref{eq:regLoss} are all non-differentiable (as the neural networks in our construction use the rectified linear unit (ReLU) activation functions), subgradient should be used in optimization, instead of gradient. 

In iteration $k$, the update rule is as follows:
\begin{equation*}
    \bm \theta^{(k+1)}\leftarrow \bm \theta^{(k)} -\alpha^{(k)}\left( -\partial \tilde{\cal L}(\bm \theta^{(k)}) + \partial \phi(\bm \theta^{(k)})\right)
\end{equation*}
where $\bm \theta =[\bm \theta_G^\top,\bm \theta_F^\top]^\top$, $\phi(\bm \theta)=\lambda_F\phi(\bm \theta_F) + \lambda_G\phi(\bm \theta_G)$ and the subgradient $\partial \tilde{\cal L} = \sum_{i=1}^M  \partial \tilde{\cal L}_i$ is computed via the chain rule and backpropagation. 

To reduce complexity, $\partial\Tilde{\cal L}$ can be approximated by sample averaging:
\[ \partial \tilde{\cal L}(\bm \theta^{(k)}) \approx \frac{1}{|{\cal B}^{(k)}|}\sum_{i\in{\cal B}^{(k)}} \partial \tilde{\cal L}_i(\bm \theta^{(k)}),\]
where ${\cal B}^{(k)}$ is a randomly sampled batch of sites such that ${\cal B}^{(k)}\subseteq [M]$.

\section{Data Simulation Details}
We simulated data to evaluate the models' ability to predict probabilities and observations as well as discover important features. 
Our data generation formula is a mixture of linear and nonlinear components. The equations below show how we generate synthetic data for each site $i$ and survey $t$. In this simulation setting, we define 10 features for both sites and surveys, and only the first five features are used to generate the responses. That is, there are five irrelevant features in each sub-model.

\begin{subequations}
\begin{align}
\mathbf{x}_{i} &\sim \mathcal{N}(
{\bm 0},\sigma^{2}{\bf I}),\\
\mathbf{w}_{it} &\sim \mathcal{N}({\bm 0},\sigma^{2}{\bf I}),\\
[\mathbf{\boldsymbol\alpha}]_k &\sim \mathcal{U}(-1,1) \mbox{ if } k = 1, \dots, 5,~[\mathbf{\boldsymbol\alpha}]_k =0,~\forall k>5,\\
[\mathbf{\boldsymbol\beta}]_{j} &\sim \mathcal{U}(-1,1) \mbox{ if } j = 1, \dots, 5,~[\mathbf{\boldsymbol\beta}]_{j}=0,~\forall j>5,\\
o_i &= \frac{\exp({(1-\rho) \cdot \boldsymbol\alpha^{T} \mathbf{x}_i + \rho \cdot \mathbf{x}_i^{T} \mathbf{A} \mathbf{x}_i})}{1+\exp({(1-\rho) \cdot \boldsymbol\alpha^{T} \mathbf{x}_i + \rho \cdot \mathbf{x}_i^{T} \mathbf{A} \mathbf{x}_i})},\\
d_{it} &= \frac{\exp( (1-\rho) \cdot \boldsymbol\beta^{T} \mathbf{w}_{it} + \rho \cdot \mathbf{w}_{it}^{T} \mathbf{B} \mathbf{w}_{it} )} {1+\exp( (1-\rho) \cdot \boldsymbol\beta^{T} \mathbf{w}_{it} + \rho \cdot \mathbf{w}_{it}^{T} \mathbf{B} \mathbf{w}_{it} )},
\end{align}
\end{subequations}

Here, $\boldsymbol\alpha$ is a coefficient vector on site features ($\mathbf{x}_{i}$) and $\boldsymbol\beta$ is a coefficient vector on survey features ($\mathbf{w}_{it}$), $\mathbf{A}$ and $\mathbf{B}$ are diagonal matrices of $\boldsymbol\alpha$ and $\boldsymbol\beta$, respectively. The value of $\rho = [0,1]$ indicates the contribution of linear and nonlinear terms in generating synthetic data. When $\rho = 0$, the latent generative models for $o_i$ and $d_{it}$ are linear models, while $\rho = 1$ corresponds to nonlinear models. We sample covariates from the normal distribution to ensure that we have well-balanced probabilities. We sampled the coefficients from the uniform distribution to avoid unbounded values.

We generated training, validation, and test sets from the same formula. We generate different types of datasets with the size of sites ($M$) and visits ($T$) and $\rho$ value. We use $M \in \{100, 1000\}$ and $T \in \{3, 10\}$ for training and validation sets and fix the site size for test sets with $M=1000$ and the corresponding value of $T$. We also generate datasets using $\rho \in \{0,1\}$. In total, we have 8 different types of datasets as described in Table ~\ref{tab:synthetic_datasets}.

\begin{table}[ht]
    \centering
    \small
    \begin{tabular}{ |c||c|c|c| } 
         \hline
         idx & nSites & nVisits & $\rho$\\ 
         \hline
         \hline
         1 & 100 & 3 & 0\\ 
         2 & 100 & 3 & 1\\ 
         3 & 100 & 10 & 0\\ 
         4 & 100 & 10 & 1\\ 
         5 & 1000 & 3 & 0\\ 
         6 & 1000 & 3 & 1\\ 
         7 & 1000 & 10 & 0\\ 
         8 & 1000 & 10 & 1\\ 
         \hline
    \end{tabular}
    \caption{Synthetic datasets}
    \label{tab:synthetic_datasets}
\end{table}

\section{Avian Point Count Dataset Details}
We also analyzed data on bird distributions to evaluate the proposed method on a real dataset. We used 10,845 5-minute point count bird surveys from the Oregon 2020 database \citep{oregon2020}. Surveys were conducted during the bird breeding season (May 15-July 10) by trained field ornithologists from 2011 to 2019. The survey locations were selected according to a stratified random design to distribute observations across Oregon. Within this design, 3-8 surveys were clustered within one randomly selected 1-square-mile section of each of Oregon's 36-square-mile township. During each survey, all birds were counted and identified to species. 

We selected five common Oregon species for this analysis. Common Yellowthroat (\textit{Geothlypis trichas}), Eurasian Collared-Dove (\textit{Streptopelia decaocto}), Song Sparrow (\textit{Melospiza melodia}), Western Meadowlark (\textit{Sturnella neglecta}), and Pacific Wren (\textit{Troglodytes pacificus}), vocalize frequently during the breeding season and have conspicuous, easily identifiable vocalizations. These species have very different habitat preferences. Common Yellowthroat is found in extremely wet vegetation with little canopy cover. Eurasian Collared-Dove is found in human-dominated habitats. Song Sparrow is more of a generalist, and is found in most habitats with rich ground-level vegetation. Western Meadowlark is found in grasslands. Pacific Wren is found in wet forests with rich undergrowth on the forest floor. 

\subsection{Environmental Features}
We compiled features for the models representing both the surrounding environment and the observation conditions.  
We constructed environmental features from a time series of radiometrically consistent, gap-free Landsat satellite image composites. 
We aggregated all summertime (Julian days 183 - 243) Landsat Collection 1 Tier 1 surface reflectance images with less than 85\% cloud cover and which intersected our study area for processing. 
We harmonized the Landsat Operational Land Imager data with the Landsat Thematic Mapper and the Landsat Enhanced Thematic Mapper plus data using the reduced major axis regression coefficients from Roy et al. \citeyear{roy2016characterization}. 
We removed clouds and cloud shadows from the imagery using the quality assessment band produced by the FMask algorithm \citep{zhu2012object,zhu2015improvement}. 
We composited each year's worth of satellite imagery into a single image using the medoid method \citep{flood2013seasonal}. 
We computed a time series of normalized burn ratio (NBR) images from the annual composites \citep{key1999normalized}. 
The LandTrendr algorithm, with the NBR time series as input, derived a time series of gap-free, fitted imagery (see \citeauthor{kennedy2015attribution} 2015 for details). 
We used Google Earth Engine \citep{gorelick2017google} for all image processing. From the time-series of fitted images 34 spectral indices were computed. Specifically, we used three components (brightness, greenness, wetness) of Tasseled Cap - TCB, TCG, TCW - and Tasseled Cap Angle (TCA) which captures the angle between the TCG and TCB values. 
\begin{table}[H]
    \centering
    \small
    \begin{tabular}{ |l|l||l|l| } 
        \hline
        ID & \multicolumn{1}{c||}{Environmental Features} & ID & \multicolumn{1}{c|}{Environmental Features}\\
        \hline
        \hline
        1 & aspect mean 75          & 15 & aspect stdDev 300\\
        2 & aspect stdDev 75        & 16 & TCA stdDev 300\\
        3 & elevation mean 75       & 17 & TCB stdDev 300\\
        4 & elevation stdDev 75     & 18 & TCW stdDev 300\\
        5 & slope stdDev 75         & 19 & aspect mean 600\\
        6 & TCA mean 75  & 20 & aspect stdDev 600\\
        7 & TCA stdDev 75 & 21 & TCB stdDev 600\\
        8 & TCB mean 75  & 22 & TCW stdDev 600\\
        9 & TCB stdDev 75 & 23 & aspect mean 1200\\
        10 & TCG stdDev 75 & 24 & aspect stdDev 1200\\
        11 & TCW stdDev 75 & 25 & TCB stdDev 1200\\
        12 & aspect stdDev 150      & 26 & TCW stdDev 1200\\
        13 & TCB stdDev 150 & 27 & aspect mean 2400\\
        14 & TCW stdDev 150 & 28 & aspect stdDev 2400\\
        \hline
    \end{tabular}
    \caption{28 environmental features used in this paper's experiments. The feature name indicates the land cover index, statistics (mean/stdDev), and radius scale.}
    \label{tab:bird_features}
\end{table}

\subsection{Observation Features}
The observation-related features were year, day, and time of observation, to capture time-varying detectability. 
In the real data experiments, the detection model had both the observation-related features and the environmental features as inputs. Even though the environmental features did not vary across surveys, they could affect detectability (e.g., vegetation affects how the sound of bird calls carries through forest). The feature selection layer of the neural networks provided a mechanism for choosing a sparser set of features.

\section{Parameter Tuning Details}
The hyper-parameters used for each model and the number and range of values tried per hyper-parameter are described in Tab.~\ref{tab:tuning_parameter_values}. The optimal values are selected based on AUPRC performance on the validation set. In this work, we assumed that the regularization weights $\lambda_F$ (for occupancy component) and $\lambda_G$ (for detection component) in \texttt{StatEcoNet} share the same value ($\lambda_F$ = $\lambda_G$ = $\lambda$). 

\begin{table}[ht!]
    \centering
    \begin{tabular}{ |l|l|l|l| } 
        \hline
        Tuning Parameter & \multicolumn{1}{c|}{\texttt{OD-LR}} & \multicolumn{1}{c|}{\texttt{OD-1NN}} & \multicolumn{1}{c|}{\texttt{StatEcoNet}}\\
        \hline
        \hline
        $learningRate$ &  \multicolumn{3}{l|}{ $\{0.0001, 0.001, 0.01\}$ } \\
        \hline
        $nEpoch$ &  \multicolumn{3}{l|}{ $[1-2000]$ for synthetic datasets,} \\
                 &  \multicolumn{3}{l|}{ $[1-1000]$ for bird datasets}\\
        \hline
        $batchSize$  &  & \multicolumn{2}{l|}{ $\{32, all\}$ }\\
        \hline
        $nNeurons$ &  & \multicolumn{2}{l|}{ $\{8, 16, 32\}$ for synthetic datasets,} \\
                   &  & \multicolumn{2}{l|}{ $\{16, 32, 64\}$ for bird datasets}\\
        \hline
        $nLayers$  & & & $\{1, 3\}$\\
        \hline
        $\ell_{2,1}$-norm weight ($\lambda$) & & & $\{0, 0.001, 0.01\}$\\
        \hline
        \hline
        \hline
        Tuning Parameter  &  \multicolumn{3}{c|}{  \texttt{OD-BRT} } \\
        \hline
        \hline
        $shrinkage$  & \multicolumn{3}{l|}{ $[0.1-1]$ }\\
        \hline
        $bagFraction$  & \multicolumn{3}{l|}{ $[0.1-1]$ }\\
        \hline
        $nTrees$  & \multicolumn{3}{l|}{ $[1-1000]$ }\\
        \hline
        $treeDepth$  & \multicolumn{3}{l|}{ $[2-10]$ }\\
        \hline
    \end{tabular}
    \caption{Tuning parameter values. For the first five rows, we explored combinations of these discrete values in a grid search. For the \texttt{OD-BRT} parameters in the bottom three rows, we explored these ranges with Bayesian optimization.}
    \label{tab:tuning_parameter_values}
\end{table}

We found that tuning the \texttt{OD-BRT} parameters was computationally intensive, so 
we selected parameters via Bayesian optimization~\citep{Snoek2012}, as implemented in the R package \texttt{rBayesianOptimization}~\citep{rBayesOptPkg}.
Since grid search evaluates every combination of the set of tuning parameters, it surely finds the best combination of those values; however, it can be inefficient to evaluate all possible combinations.
In contrast, Bayesian optimization searches for parameter values in a range, potentially evaluating parameter values beyond the fixed values used in grid search. 
This allows for the possibility of finding better combinations of parameter values than those specified by grid search, though it may not always find the optimal values among all possibilities.
We found that the Bayesian optimization method found tuning parameter values with higher AUPRC than grid search in less time.

\section{Simulation Results}
\subsection{Linear Latent Model ($\rho=0$)}
\subsubsection{Optimal parameters}
\begin{table}[H]
    \centering
    \begin{tabular}{ |c|c|c|c|c|c| } 
        \hline
        \multirow{2}{*}{Model} & \multirow{2}{*}{Hyper-parameter} & \multicolumn{4}{c|}{Optimal Values} \\
        \cline{3-6}
         &  & 100x3 & 100x10 & 1000x3 & 1000x10\\
        \hline
        \texttt{OD-LR} & $learningRate$ & 0.01 & 0.01 & 0.01 & 0.01 \\
        \hline
        \texttt{OD-1NN} & $learningRate$ & 0.001 & 0.001 & 0.001 & 0.001\\
          & $batchSize$ & 32 & 32 & 32 & 32\\
          & $nNeurons$ & 16 & 16 & 16 & 32\\
        \hline
        \texttt{StatEcoNet} & $learningRate$ & 0.001 & 0.001 & 0.001 & 0.001\\
          & $batchSize$ & $all$ & 32 & 32 & 32\\
          & $nNeurons$ & 8 & 8 & 8 & 8\\
          & $nLayers$ & 1 & 3 & 1 & 3\\
          & $\lambda$ & 0 & 0.01 & 0.01 & 0.01\\
        \hline
         \texttt{OD-BRT} & $shrinkage$ & 0.2399  & 0.3629 & 0.1407 & 0.123\\
         & $bagFraction$ & 0.6279 & 0.5107 &  0.8759 & 0.4817\\
          & $treeDepth$ & 2 & 2 & 4 & 6\\
        \hline
    \end{tabular}
    \caption{Optimal parameters for linear latent models.}
    \label{tab:opt_parameters_linear}
\end{table}

\subsubsection{Predictive performance}
\begin{figure}[H]
    \centering
    \begin{subfigure}[t]{.4\textwidth}
        \centering
        \includegraphics[width=\linewidth]{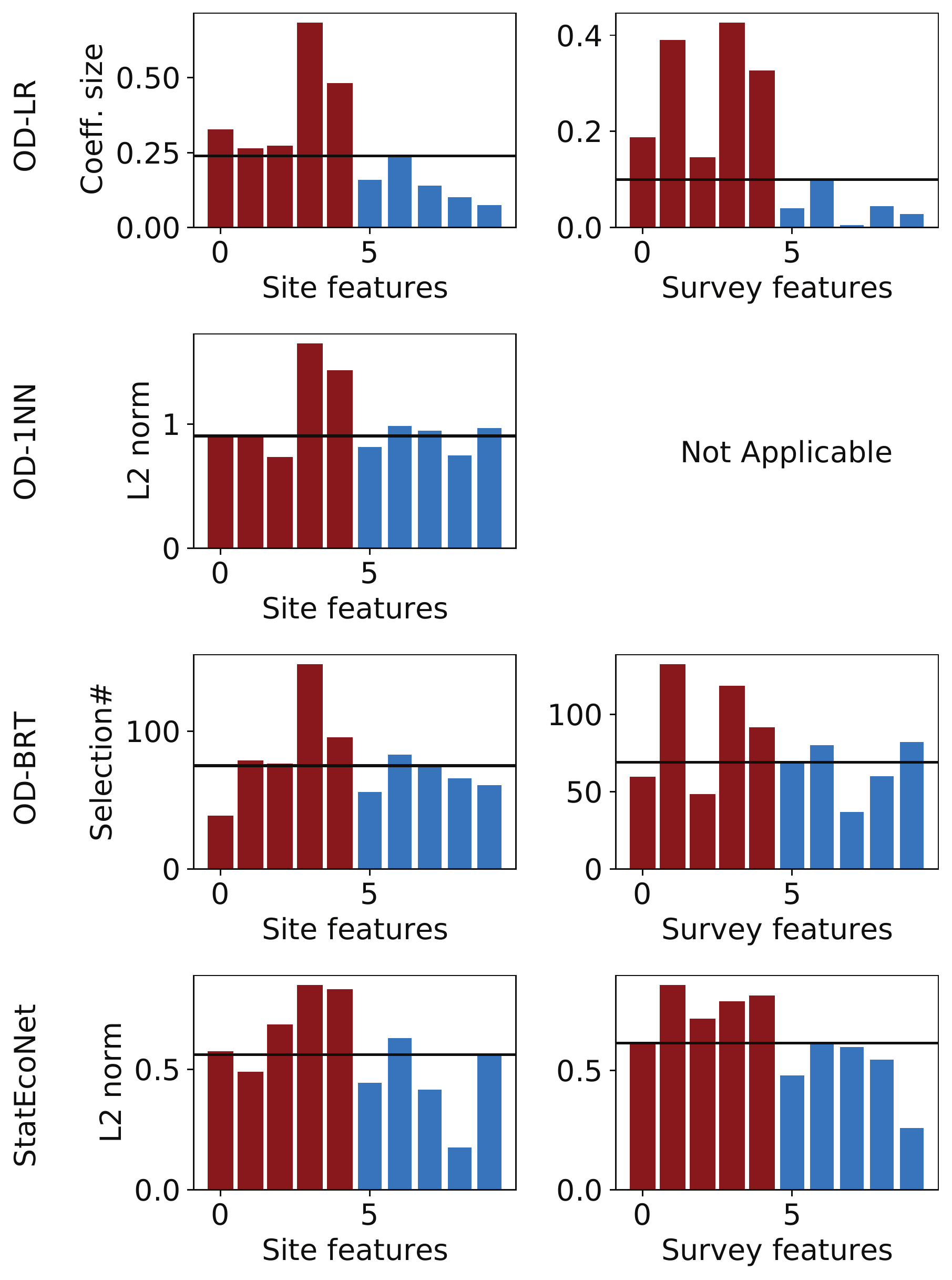}
        \caption{$M$=100, $T$=3 ($\lambda$=0)}\label{fig:fig_1200x3x0x1}
    \end{subfigure}
    \qquad
    \begin{subfigure}[t]{.4\textwidth}
        \centering
        \includegraphics[width=\linewidth]{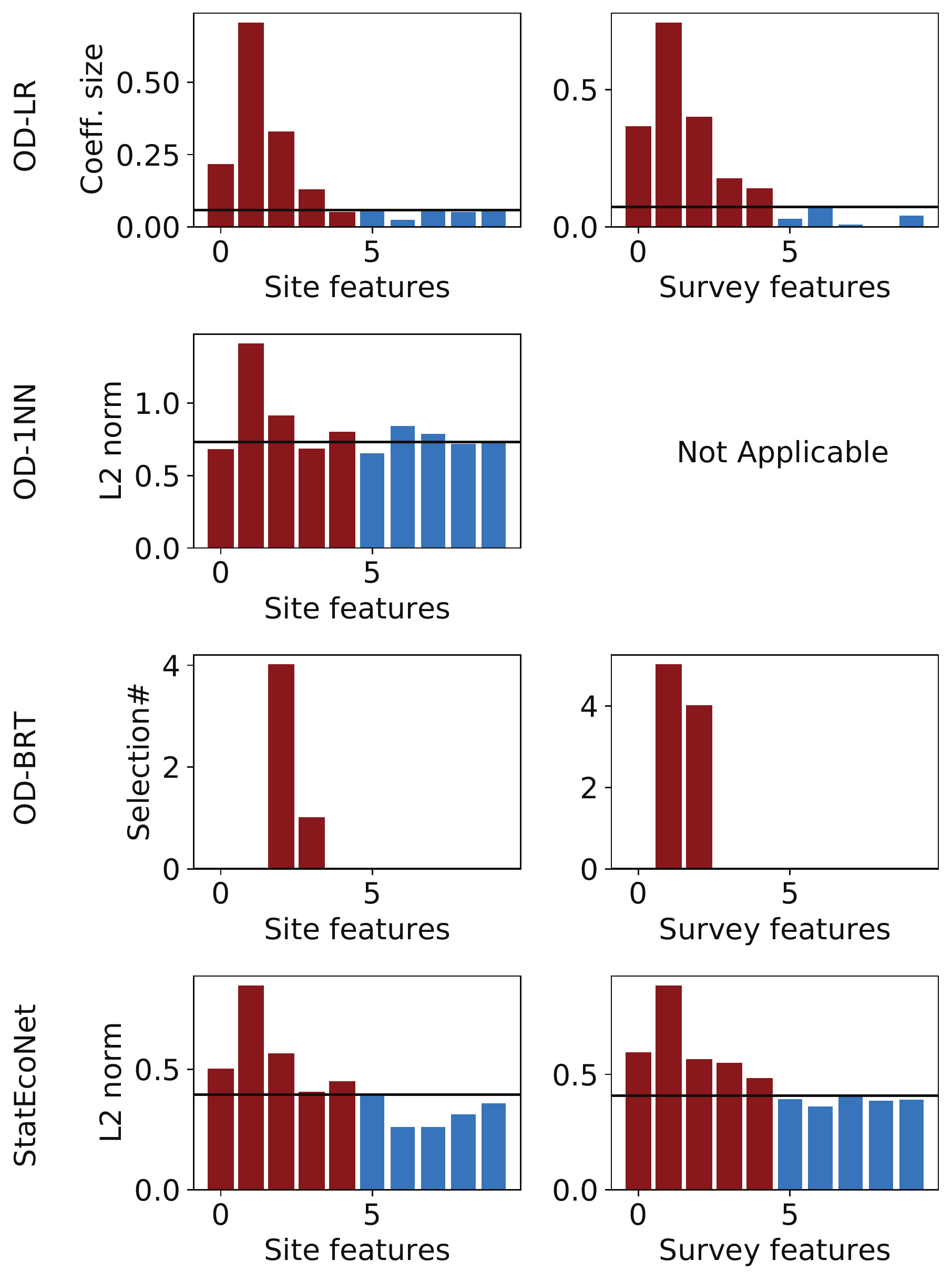}
        \caption{$M$=100, $T$=10 ($\lambda$=0.01)}\label{fig:fig_1200x10x0x1}
    \end{subfigure}
    \medskip
    \begin{subfigure}[t]{.4\textwidth}
        \centering
        \includegraphics[width=\linewidth]{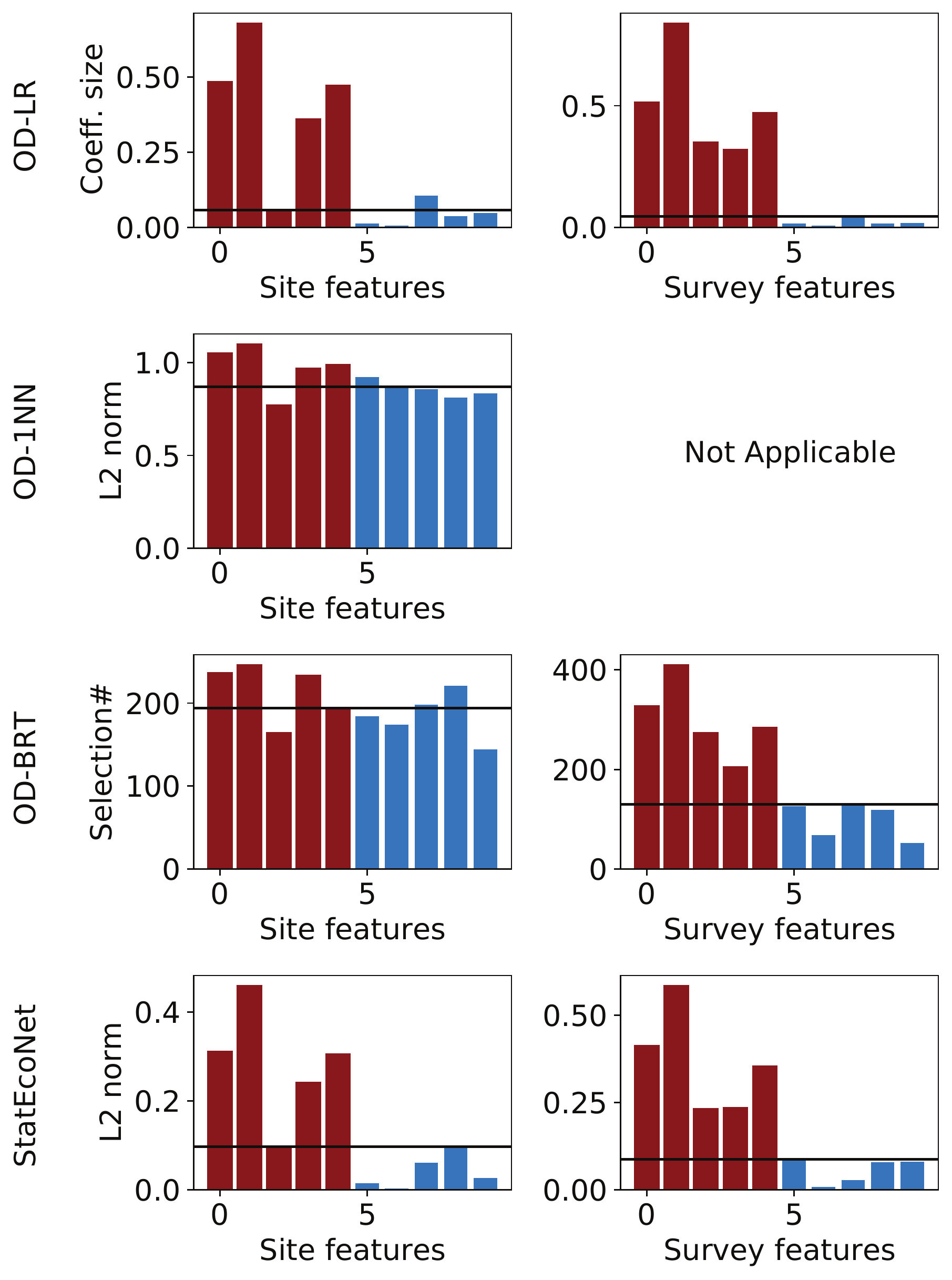}
        \caption{$M$=1000, $T$=3 ($\lambda$=0.01)}\label{fig:fig_3000x3x0x1}
    \end{subfigure}
    \qquad
    \begin{subfigure}[t]{.4\textwidth}
        \centering
        \includegraphics[width=\linewidth]{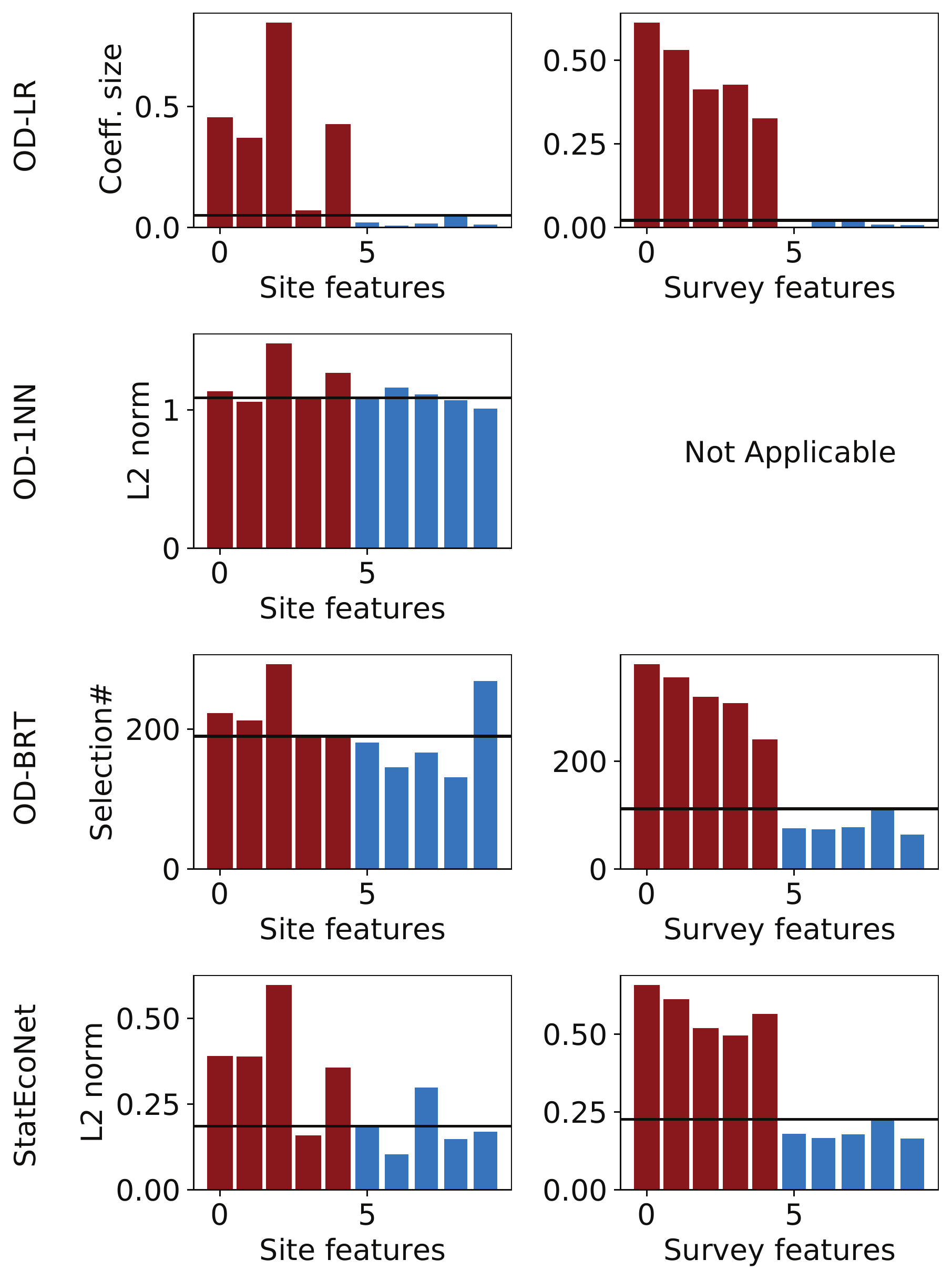}
        \caption{$M$=1000, $T$=10 ($\lambda$=0.01)}\label{fig:fig_3000x10x0x1}
    \end{subfigure}
\caption{Selected features by each method for the synthetic dataset with linear relationships. The red bars correspond to relevant features, and the blue bars irrelevant features. $M$ is the number of training sites and $T$ is the number of visits per site. $\lambda$ is the optimal regularization weights for $\lambda_F$ and $\lambda_G$. The second plot of \texttt{OD-1NN} is not available here because survey features are combined with outputs of a hidden layer from that method. The horizontal black line indicates the top 5 features according to the importance scores (y-axis).}
\label{fig:syn_feature_linear}
\end{figure}

Our model comparisons on simulated data with linear feature combinations indicates that the linear model, \texttt{OD-LR}, performs best on linear data, as expected. However, it is rare that all feature relationships would be linear \textit{and} that the modeler would know this in advance. Considering the more general case with unknown feature relationships, the results show that \texttt{StatEcoNet} performs similarly to \texttt{OD-LR} for recovering the true model probabilities (Tab.~\ref{tab:linear_test_performance} correlation columns), predicting new data (Tab.~\ref{tab:linear_test_performance} AUPRC and AUROC columns), and selecting the correct features (Fig.~\ref{fig:fig_3000x3x0x1}). 
The \texttt{OD-1NN} and \texttt{OD-BRT} models exhibit problems on some datasets, notably with detection probability correlations (Tab.~\ref{tab:linear_test_performance}) and occupancy feature selection (Fig.~\ref{fig:fig_1200x10x0x1}).

\begin{table}[ht!]
    \centering
    \footnotesize
    \begin{tabular}{ |c|c|c|c|c|c|c| } 
        \hline
        Data size & Method & Training Time & Occ.Prob.Corr. & Det.Prob.Corr. & AUPRC & AUROC \\ 
        \hline
        \hline
         & \texttt{OD-LR}      & 4.58 $\pm$ 3.62           &\textbf{0.91} $\pm$ 0.03 & \textbf{0.96} $\pm$ 0.02  & \textbf{0.63} $\pm$ 0.004 & \textbf{0.84} $\pm$ 0.002 \\
        $M=100$ & \texttt{OD-1NN}     & 9.87 $\pm$ 2.54           &  0.86 $\pm$ 0.03          & 0.82 $\pm$ 0.02           & 0.59 $\pm$ 0.01           & 0.81 $\pm$ 0.004 \\
        $T=3$ & \texttt{OD-BRT}     & \textbf{3.44} $\pm$ 2.47  &  0.78 $\pm$ 0.02          & 0.84 $\pm$ 0.02           & 0.57 $\pm$ 0.01           & 0.80 $\pm$ 0.01 \\
        & \texttt{StatEcoNet} & 11.29 $\pm$ 3.62          &  0.87 $\pm$ 0.02          & 0.95 $\pm$ 0.01           & 0.62 $\pm$ 0.01           & 0.83 $\pm$ 0.01 \\
        \hline
        \hline
         & \texttt{OD-LR}      &  6.16 $\pm$ 3.53          & \textbf{0.93} $\pm$ 0.01              & \textbf{0.98} $\pm$ 0.01  &  \textbf{0.71} $\pm$ 0.003    & \textbf{0.87} $\pm$ 0.002 \\
        $M=100$ & \texttt{OD-1NN}     &  10.10 $\pm$ 5.78         & 0.92 $\pm$ 0.03                       & 0.93 $\pm$ 0.01           & 0.67 $\pm$ 0.01               & 0.85 $\pm$ 0.01 \\
        $T=10$& \texttt{OD-BRT}     &  \textbf{2.66} $\pm$ 4.07 & 0.81 $\pm$ 0.03                       & 0.88 $\pm$ 0.05           & 0.62 $\pm$ 0.03               & 0.81 $\pm$ 0.02 \\
        & \texttt{StatEcoNet} &  3.51 $\pm$ 0.85          & \textbf{0.93} $\pm$ 0.02              & 0.97 $\pm$ 0.01           & 0.70 $\pm$ 0.01               & \textbf{0.87} $\pm$ 0.004 \\
        \hline
        \hline
         & \texttt{OD-LR}      &  20.65 $\pm$ 7.60         & \textbf{0.99} $\pm$ 0.0001    & \textbf{1.00} $\pm$ 0.0002    & \textbf{0.68} $\pm$ 0.0003    & \textbf{0.86} $\pm$ 0.0001 \\
         $M=1000$& \texttt{OD-1NN}     &  6.75 $\pm$ 0.67          & 0.98 $\pm$ 0.002              & 0.98 $\pm$ 0.004              & 0.66 $\pm$ 0.004              & \textbf{0.86} $\pm$ 0.001 \\
        $T=3$ & \texttt{OD-BRT}     &  \textbf{1.19} $\pm$ 0.85 & 0.77 $\pm$ 0.05               & 0.75 $\pm$ 0.02               & 0.53 $\pm$ 0.03               & 0.76 $\pm$ 0.02 \\
        & \texttt{StatEcoNet} &  9.79 $\pm$ 5.57          & 0.98 $\pm$ 0.002              & \textbf{1.00} $\pm$ 0.001     & \textbf{0.68} $\pm$ 0.001     & \textbf{0.86} $\pm$ 0.001 \\
        \hline
        \hline
        &\texttt{OD-LR}      &  12.82 $\pm$ 8.33         & \textbf{0.99} $\pm$ 0.003     & \textbf{1.00} $\pm$ 0.0004    & \textbf{0.68} $\pm$ 0.001    & \textbf{0.87} $\pm$ 0.001 \\
        $M=1000$ &\texttt{OD-1NN}     &  6.06 $\pm$ 1.21          & 0.99 $\pm$ 0.002              & 0.99 $\pm$ 0.001              & 0.67 $\pm$ 0.001              & 0.86 $\pm$ 0.001 \\
        $T=10$ &\texttt{OD-BRT}     &  5.88 $\pm$ 1.23          & 0.86 $\pm$ 0.02      & 0.79 $\pm$ 0.01               & 0.53 $\pm$ 0.01               & 0.78 $\pm$ 0.01 \\
        &\texttt{StatEcoNet} &  \textbf{5.04} $\pm$ 2.34 & \textbf{0.99} $\pm$ 0.003     & 0.99 $\pm$ 0.001              & \textbf{0.68} $\pm$ 0.002     & 0.86 $\pm$ 0.001 \\
        \hline
    \end{tabular}
    \caption{Performance metrics (mean $\pm$ st. dev.) on simulated data with linear relationships.}
    \label{tab:linear_test_performance}
\end{table}

\clearpage

\subsection{Nonlinear Latent Model ($\rho=1$)}
\subsubsection{Optimal parameters}
\begin{table}[H]
    \centering
    \begin{tabular}{ |c|c|c|c|c|c| } 
        \hline
        \multirow{2}{*}{Model} & \multirow{2}{*}{Hyper-parameter} & \multicolumn{4}{c|}{Optimal Values} \\
        \cline{3-6}
         &  & 100x3 & 100x10 & 1000x3 & 1000x10\\
        \hline
        \texttt{OD-LR} & $learningRate$ & 0.01 & 0.01 & 0.01 & 0.01 \\
        \hline
        \texttt{OD-1NN} & $learningRate$ & 0.001 & 0.001 & 0.001 & 0.001\\
          & $batchSize$ & $all$ & $all$ & 32 & 32\\
          & $nNeurons$ & 16 & 16 & 32 & 8\\
        \hline
        \texttt{StatEcoNet} & $learningRate$ & 0.001 & 0.001 & 0.001 & 0.001\\
          & $batchSize$ & 32 & 32 & $all$ & $all$\\
          & $nNeurons$ & 32 & 8 & 16 & 16\\
          & $nLayers$ & 3 & 1 & 3 & 3\\
          & $\lambda$ & 0.001 & 0.001 & 0.01 & 0.01\\
        \hline
         \texttt{OD-BRT} & $shrinkage$ & 0.9440  & 0.3320 & 0.5149 & 0.4040\\
         & $bagFraction$ & 0.1435 & 0.6444 &  0.7826 & 0.7499\\
          & $treeDepth$ & 5 & 9 & 2 & 3\\
        \hline
    \end{tabular}
    \caption{Optimal parameters for nonlinear latent models.}
    \label{tab:opt_parameters_nonlinear}
\end{table}

\subsubsection{Predictive performance}
On the simulation experiments where the data generation uses nonlinear feature combinations, \texttt{StatEcoNet} performs well. On only the smallest datasets ($M=100$), it is outperformed by \texttt{OD-BRT} in terms of recovering the occupancy and detection probabilities as well as predicting new data (Tab.~\ref{tab:nonlinear_test_performance}). On the larger datasets ($M=1000$), \texttt{StatEcoNet} performs as well or better than \texttt{OD-BRT}, and the training time starts to favor \texttt{StatEcoNet} heavily as dataset sizes increase. On these larger datasets, \texttt{StatEcoNet} also has an advantage for feature selection.

\begin{table}[ht!]
    \centering
    \footnotesize
    \begin{tabular}{ |c|c|c|c|c|c|c| } 
        \hline
        Data size & Method & Training Time & Occ.Prob.Corr. & Det.Prob.Corr. & AUPRC & AUROC \\ 
        \hline
        \hline
        & \texttt{OD-LR}      & \textbf{0.40} $\pm$ 0.50  & -0.002 $\pm$ 0.02         & 0.003 $\pm$ 0.01              & 0.29 $\pm$ 0.004          &  0.50 $\pm$  0.004 \\
        $M=100$ &\texttt{OD-1NN}     & 5.77 $\pm$ 12.33          & 0.05 $\pm$ 0.07           & -0.01 $\pm$ 0.02              & 0.29 $\pm$ 0.01           & 0.50 $\pm$ 0.02\\
        $T=3$ &\texttt{OD-BRT}     & 1.48 $\pm$ 1.07           & \textbf{0.38} $\pm$0.11   & \textbf{0.55} $\pm$0.15       & \textbf{0.37} $\pm$ 0.03  & \textbf{0.60} $\pm$ 0.02\\
        &\texttt{StatEcoNet} & 4.91 $\pm$ 6.85           & 0.1 $\pm$ 0.12            & 0.16 $\pm$ 0.21               & 0.31 $\pm$ 0.03           & 0.53 $\pm$ 0.04\\
        \hline
        \hline
        & \texttt{OD-LR}      &  \textbf{0.90} $\pm$ 0.52 & -0.003 $\pm$ 0.02         & 0.01 $\pm$ 0.004           & 0.39 $\pm$ 0.01              & 0.51 $\pm$ 0.01 \\
        $M=100$ & \texttt{OD-1NN}     &  16.13 $\pm$ 16.08        & 0.11 $\pm$ 0.15           & 0. $\pm$ 0.01              & 0.39 $\pm$ 0.01              & 0.52 $\pm$ 0.01 \\
        $T=10$ & \texttt{OD-BRT}     &  8.07 $\pm$ 0.88          & \textbf{0.59} $\pm$ 0.01  & \textbf{0.80} $\pm$ 0.01   & \textbf{0.53} $\pm$ 0.002    & \textbf{0.66} $\pm$ 0.01 \\
        & \texttt{StatEcoNet} &  39.57 $\pm$ 50.71        & 0.03 $\pm$ 0.08           & 0.31 $\pm$ 0.38            & 0.42 $\pm$ 0.06              & 0.55 $\pm$ 0.07 \\
        \hline
        \hline
        & \texttt{OD-LR}      &  \textbf{1.21} $\pm$ 1.21 & -0.02 $\pm$ 0.04         & -0.01 $\pm$ 0.03           & 0.35 $\pm$ 0.01              & 0.50 $\pm$ 0.01 \\
        $M=1000$ & \texttt{OD-1NN}     &  18.96 $\pm$ 2.47        & 0.73 $\pm$ 0.02           & -0.02 $\pm$ 0.01              & 0.42 $\pm$ 0.01              & 0.59 $\pm$ 0.01 \\
        $T=3$ & \texttt{OD-BRT}     &  28.77 $\pm$ 22.9          & \textbf{0.79} $\pm$ 0.03  & 0.88 $\pm$ 0.04   & \textbf{0.55} $\pm$ 0.01    & \textbf{0.70} $\pm$ 0.01 \\
        & \texttt{StatEcoNet} &  25.85 $\pm$ 14.16        & 0.54 $\pm$ 0.04           & \textbf{0.90} $\pm$ 0.03            & 0.53 $\pm$ 0.02              & \textbf{0.70} $\pm$ 0.01 \\
        \hline
        \hline
        & \texttt{OD-LR}      &  \textbf{3.66} $\pm$ 3.11 s            & 0.05 $\pm$ 0.001          & 0.01 $\pm$ 0.001          & 0.32 $\pm$ 0.002          & 0.51 $\pm$ 0.001\\
        $M=1000$ & \texttt{OD-1NN}     & 30.3 $\pm$ 5.15 s    & \textbf{0.84} $\pm$ 0.01  & 0.004 $\pm$ 0.003         & 0.39 $\pm$ 0.004          & 0.61 $\pm$ 0.01 \\
        $T=10$ & \texttt{OD-BRT}     & 320 $\pm$ 60.6 s              & 0.83 $\pm$ 0.01           & \textbf{0.97} $\pm$ 0.002 & \textbf{0.53} $\pm$ 0.003 & 0.72 $\pm$ 0.002\\
        & \texttt{StatEcoNet} & 94.2 $\pm$ 17.5 s             & \textbf{0.84} $\pm$ 0.01  & \textbf{0.97} $\pm$ 0.003 & \textbf{0.53} $\pm$ 0.001 & \textbf{0.73} $\pm$ 0.003\\
        \hline
    \end{tabular}
    \caption{Performance metrics (mean $\pm$ st. dev.) on simulated data with nonlinear relationships.}
    \label{tab:nonlinear_test_performance}
\end{table}

\clearpage

\begin{figure}
    \centering
    \begin{subfigure}[t]{.4\textwidth}
        \centering
        \includegraphics[width=\linewidth]{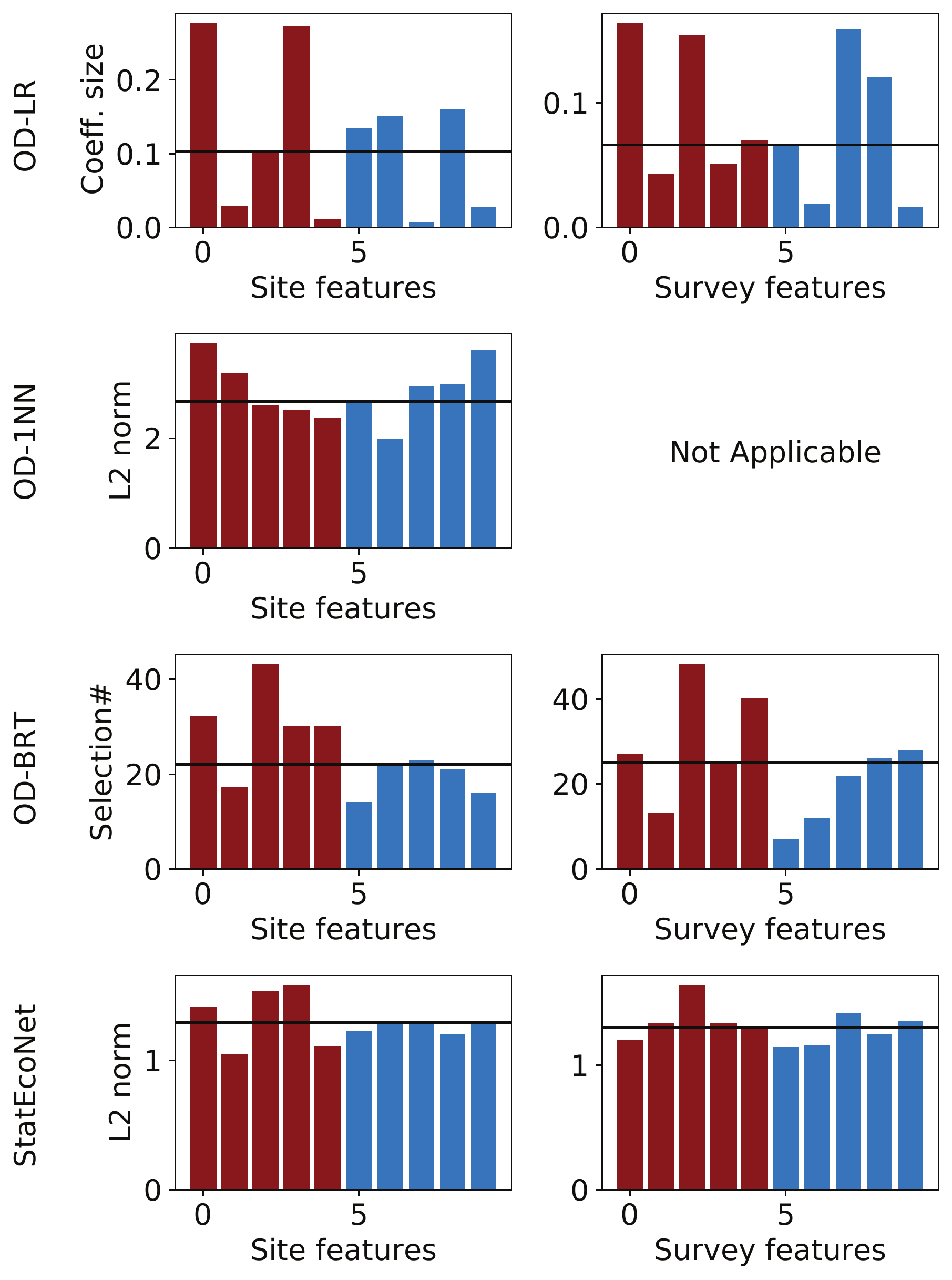}
        \caption{$M$=100, $T$=3 ($\lambda$=0.001)}\label{fig:fig_1200x3x1x1}
    \end{subfigure}
    \qquad
    \begin{subfigure}[t]{.4\textwidth}
        \centering
        \includegraphics[width=\linewidth]{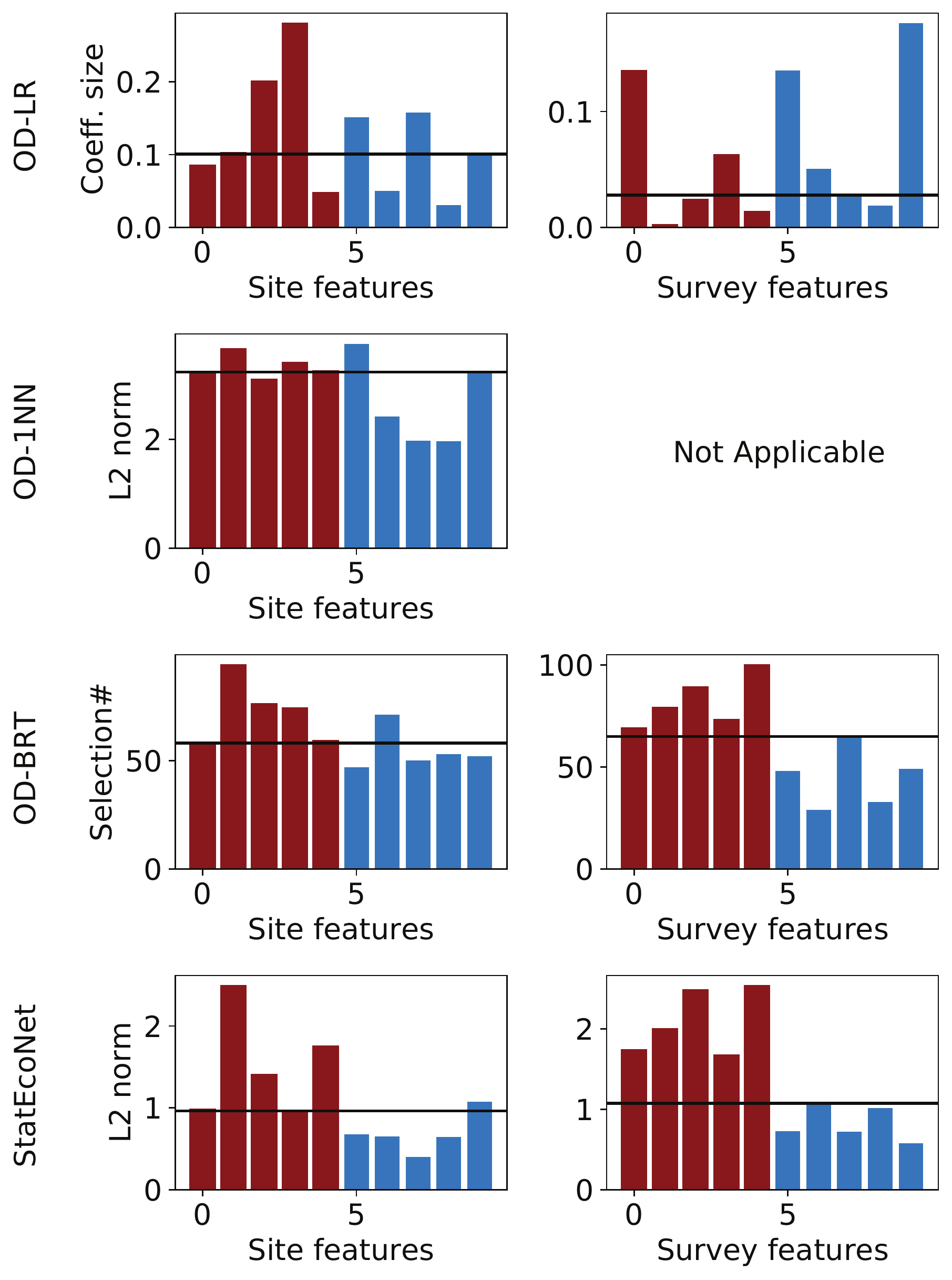}
        \caption{$M$=100, $T$=10 ($\lambda$=0.001)}\label{fig:fig_1200x10x1x1}
    \end{subfigure}
    \medskip
    \begin{subfigure}[t]{.4\textwidth}
        \centering
        \includegraphics[width=\linewidth]{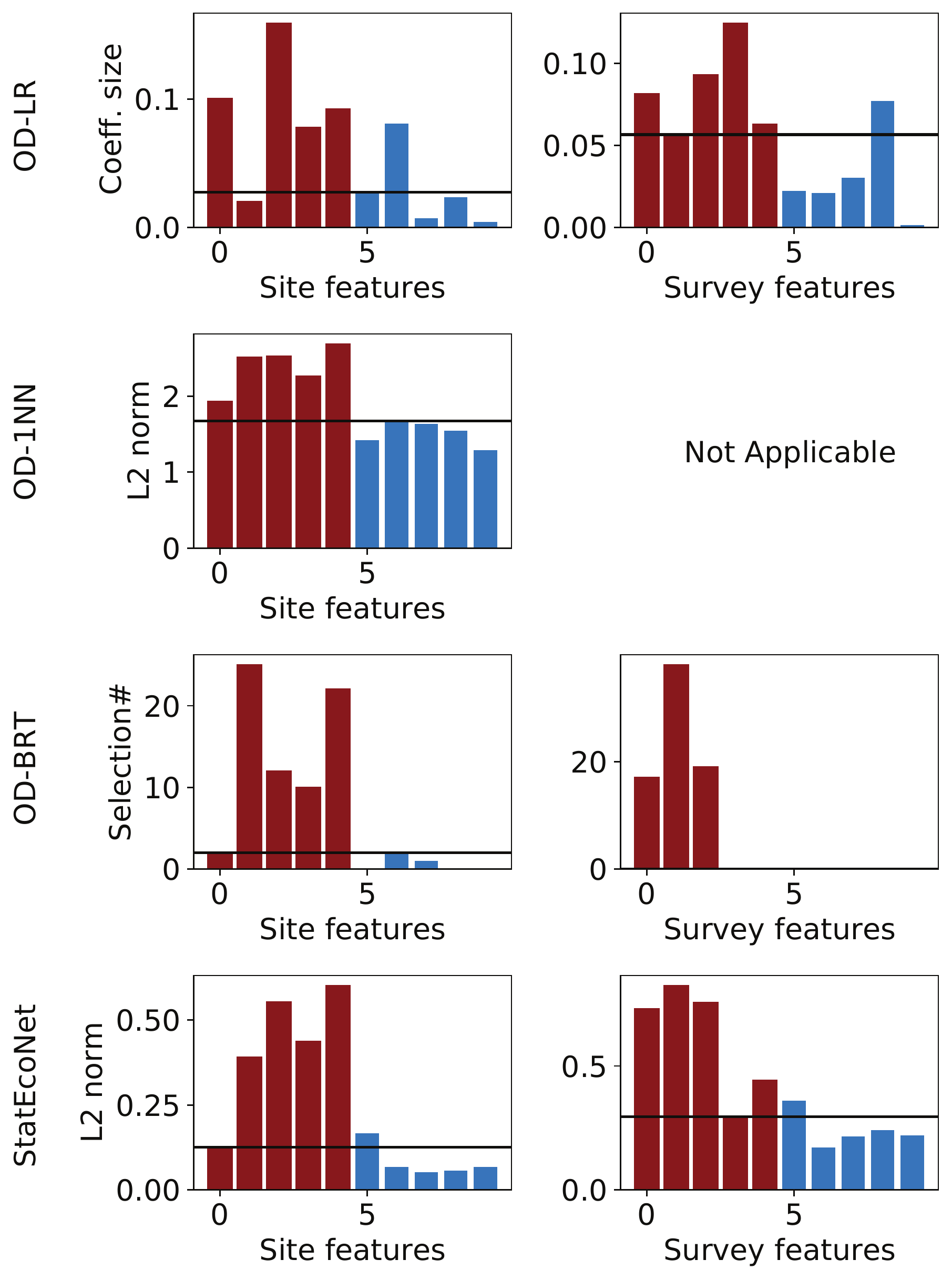}
        \caption{$M$=1000, $T$=3 ($\lambda$=0.01)}\label{fig:fig_3000x3x1x1}
    \end{subfigure}
    \qquad
    \begin{subfigure}[t]{.4\textwidth}
        \centering
        \includegraphics[width=\linewidth]{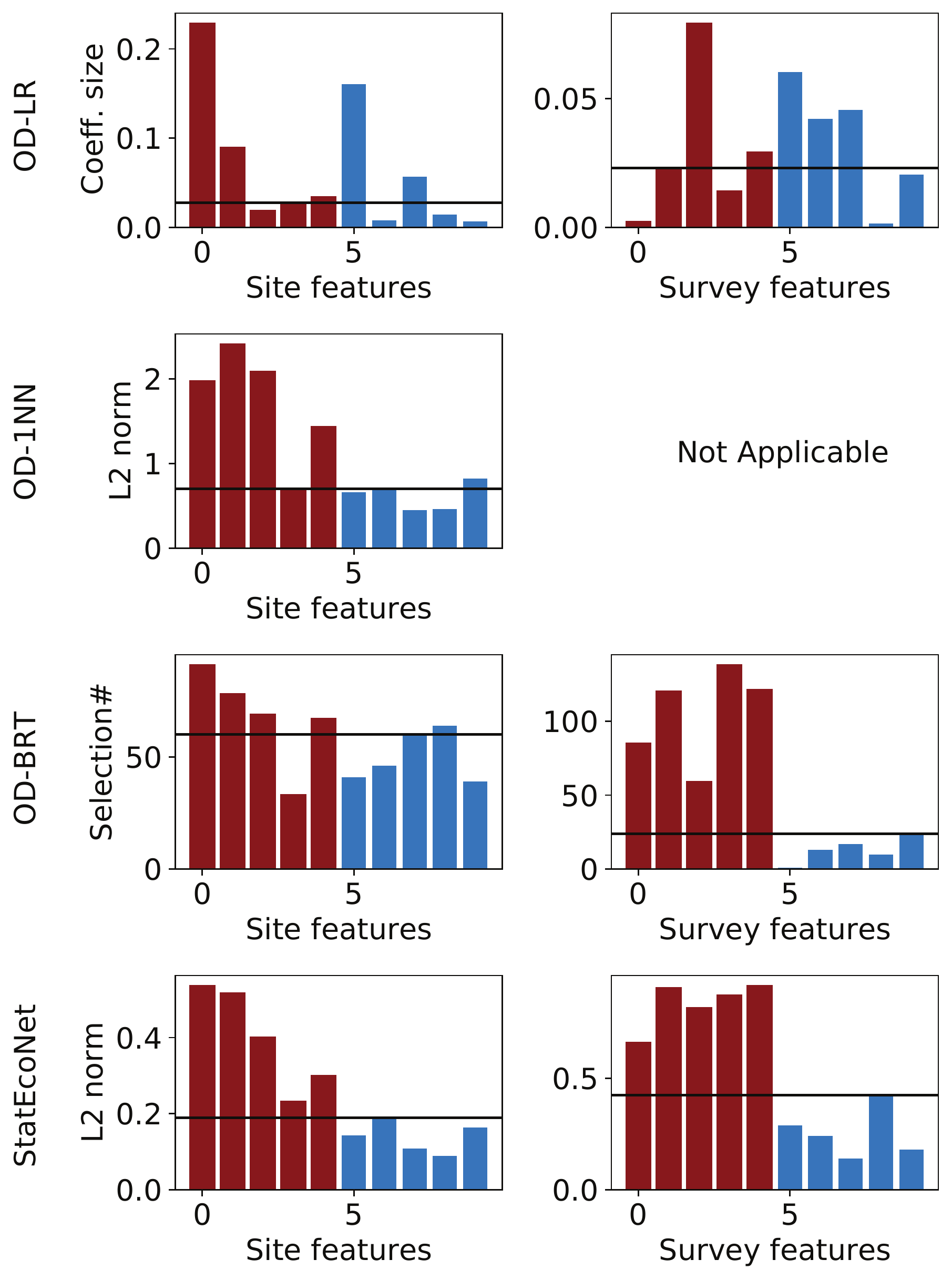}
        \caption{$M$=1000, $T$=10 ($\lambda$=0.01)}\label{fig:fig_3000x10x1x1}
    \end{subfigure}
    \caption{Selected features by each method for the synthetic dataset with nonlinear relationships. The red bars correspond to relevant features, and the blue bars irrelevant features. $M$ is the number of training sites and $T$ is the number of visits per site. $\lambda$ is the optimal regularization weights for $\lambda_F$ and $\lambda_G$. The second plot of \texttt{OD-1NN} is not available here because survey features are combined with outputs of a hidden layer from that method. The horizontal black line indicates the top 5 features according to the importance scores.}
    \label{fig:syn_feature_nonlinear}
\end{figure}

\clearpage

\section{Avian Point Count Results}\label{sec:realdata}
Each subsection below reports more detailed results for each of the five species. Each section gives a histogram of the learned occupancy and detection probabilities, feature importances for the occupancy and detection models across methods, and the optimal hyperparameters that resulted from the tuning process.

There are a few overall trends in the avian point count study results to point out. First, the differences between methods in terms of AUPRC and AUROC are minor (Table \ref{tab:AUCs}), even though the interpretations of the learned models vary substantially (visualizations in species-specific subsections below). (The results of \texttt{OD-1NN} and \texttt{StatEcoNet} have been updated from the main paper with some small changes to the parameter tuning; the main trends are unchanged.) Second, \texttt{OD-BRT} sometimes produces probability histograms that are concentrated around 0.5. These seem unrealistic and appear underfit, despite careful parameter tuning. Third, note that the detection feature importance plots are missing for \texttt{OD-1NN} for all species because this inference is not available from that method due to the architecture of the neural network.

There are also a few things to consider when viewing the probability histograms below. First, the upper left corners should be interpreted loosely. As the occupancy probability for a given point approaches zero, the contribution of the detection model for that point gets less influence in the likelihood function. Second, variation in detection probability is often biologically plausible (with some exceptions). Finally, for specialists, a low or bimodal distribution of occupancy probabilities for the non-detections makes sense, since sites will be obviously suitable or unsuitable, and some suitable sites may have non-detections.

\begin{table}[H]
    \centering
    \begin{tabular}{ |c|c|c|c|c|c| } 
        \hline
        Species & Metric & \texttt{OD-LR} & \texttt{OD-1NN} & \texttt{OD-BRT} &  \texttt{StatEcoNet}\\
        \hline
        \hline
        COYE & AUPRC & 0.375 $\pm$ 0.0614 & 0.376 $\pm$ 0.0495 & 0.369 $\pm$ 0.0458 & \textbf{0.383} $\pm$ 0.0519\\
        \hline
        EUCD & AUPRC &0.208 $\pm$ 0.0462 & 0.272 $\pm$ 0.0462 & 0.183 $\pm$ 0.0453 & \textbf{0.283} $\pm$ 0.0610\\
        \hline
         SOSP & AUPRC & 0.563 $\pm$ 0.0230 & 0.567 $\pm$ 0.0311 & 0.558 $\pm$ 0.0322 & \textbf{0.571} $\pm$ 0.021\\
        \hline
        WEME & AUPRC & 0.559 $\pm$ 0.132 & 0.545 $\pm$ 0.1269 & \textbf{0.634} $\pm$ 0.0665 & 0.593$\pm$0.1049\\
        \hline
        PAWR & AUPRC & 0.474 $\pm$ 0.0382 & 0.461 $\pm$ 0.0311 & 0.473 $\pm$ 0.0348 & \textbf{0.496} $\pm$ 0.0314\\
        \hline
        \hline
        COYE & AUROC & 0.834 $\pm$ 0.0355 & \textbf{0.836} $\pm$ 0.0229 & 0.834 $\pm$ 0.0404 & 0.828 $\pm$ 0.0375\\
        \hline
        EUCD & AUROC & 0.756 $\pm$ 0.0325 & \textbf{0.809} $\pm$ 0.03 & 0.72 $\pm$ 0.0709 & \textbf{0.809} $\pm$ 0.021\\
        \hline
        SOSP & AUROC & 0.797 $\pm$ 0.0175 & 0.801 $\pm$ 0.0192 & 0.802 $\pm$ 0.0185 & \textbf{0.803} $\pm$ 0.0152\\
        \hline
        WEME & AUROC & 0.881 $\pm$ 0.0516 &  0.891 $\pm$ 0.0416 & \textbf{0.912} $\pm$ 0.0283 & 0.910 $\pm$ 0.0292\\
        \hline
        PAWR & AUROC & 0.858 $\pm$ 0.0178 & 0.865 $\pm$ 0.0218 & 0.868 $\pm$ 0.0309 & \textbf{0.875} $\pm$ 0.026 \\
        \hline
    \end{tabular}
    \caption{Predictive performance of methods for five bird species.}
    \label{tab:AUCs}
\end{table}

\clearpage


\subsection{Common Yellowthroat (COYE)}
Common Yellowthroat (COYE) is found in extremely wet vegetation with little canopy cover. Like all songbirds, it sings more in the early morning than later in the day, so it is more frequently detected on early surveys.

\begin{figure}[H]
\centering
  \includegraphics[width=\linewidth]{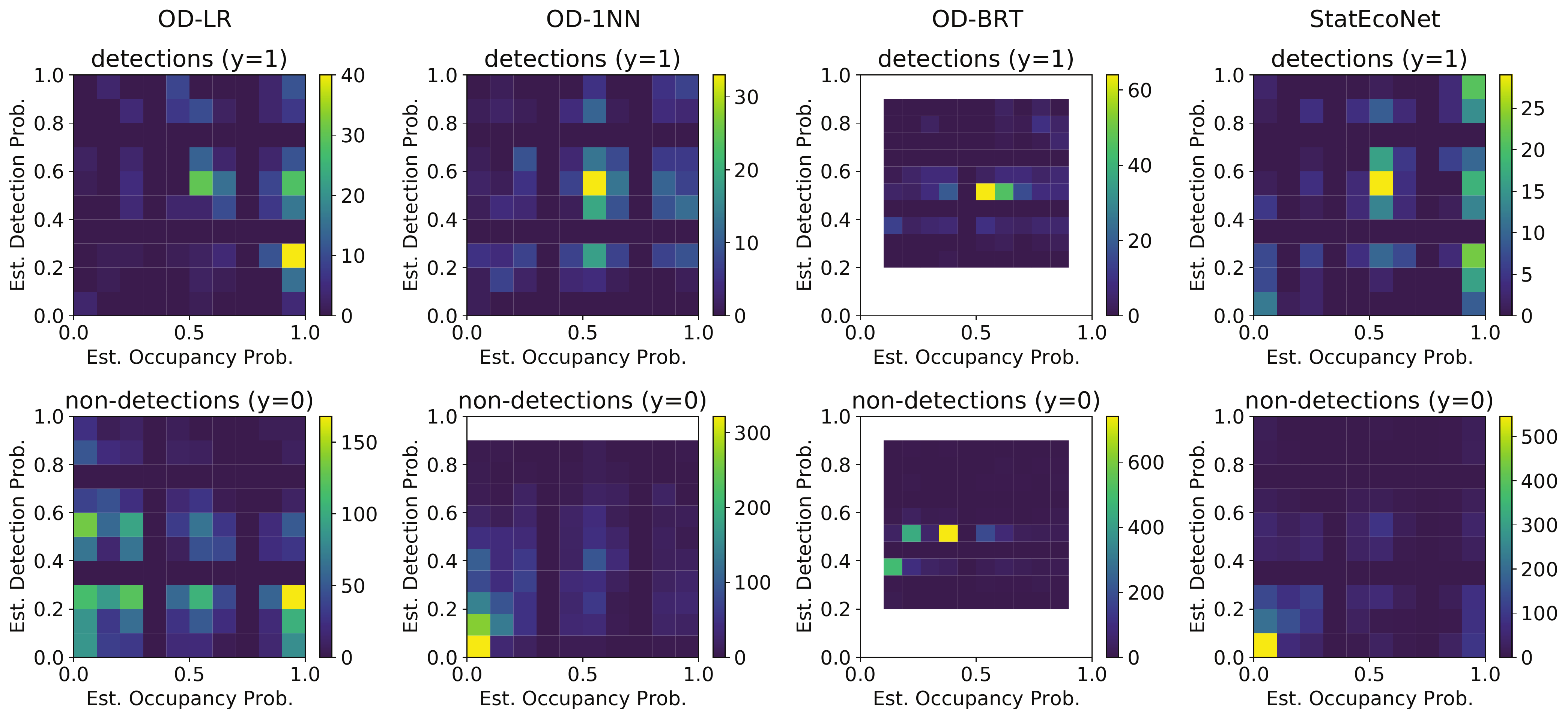}
  \caption{Histograms for Common Yellowthroat. The top row shows the occupancy and detection probability histograms for positives (detections, $y=1$), and the bottom row shows the same for negatives (non-detections, $y=0$). }
  \label{fig:COYE_densities}
\end{figure}

Figure~\ref{fig:COYE_densities} shows two-dimensional histograms of the occupancy and detection probabilities for all positive species reports (detections, $y=1$) in the top row, and all negative species reports (non-detections, $y=0$) in the bottom row. The concentration of negatives in the lower left corner of the histograms of \texttt{StatEcoNet} and \texttt{OD-1NN} may reflect the fact that much of the surveyed points are not suitable habitat for this species, so many occupancy probabilities should be low. In contrast, \texttt{OD-LR} is less believable, with many negatives having high occupancy probability, implying that the species was missed more frequently than is realistic.

\clearpage 

\begin{figure}[H]
\centering
  \includegraphics[width=\linewidth]{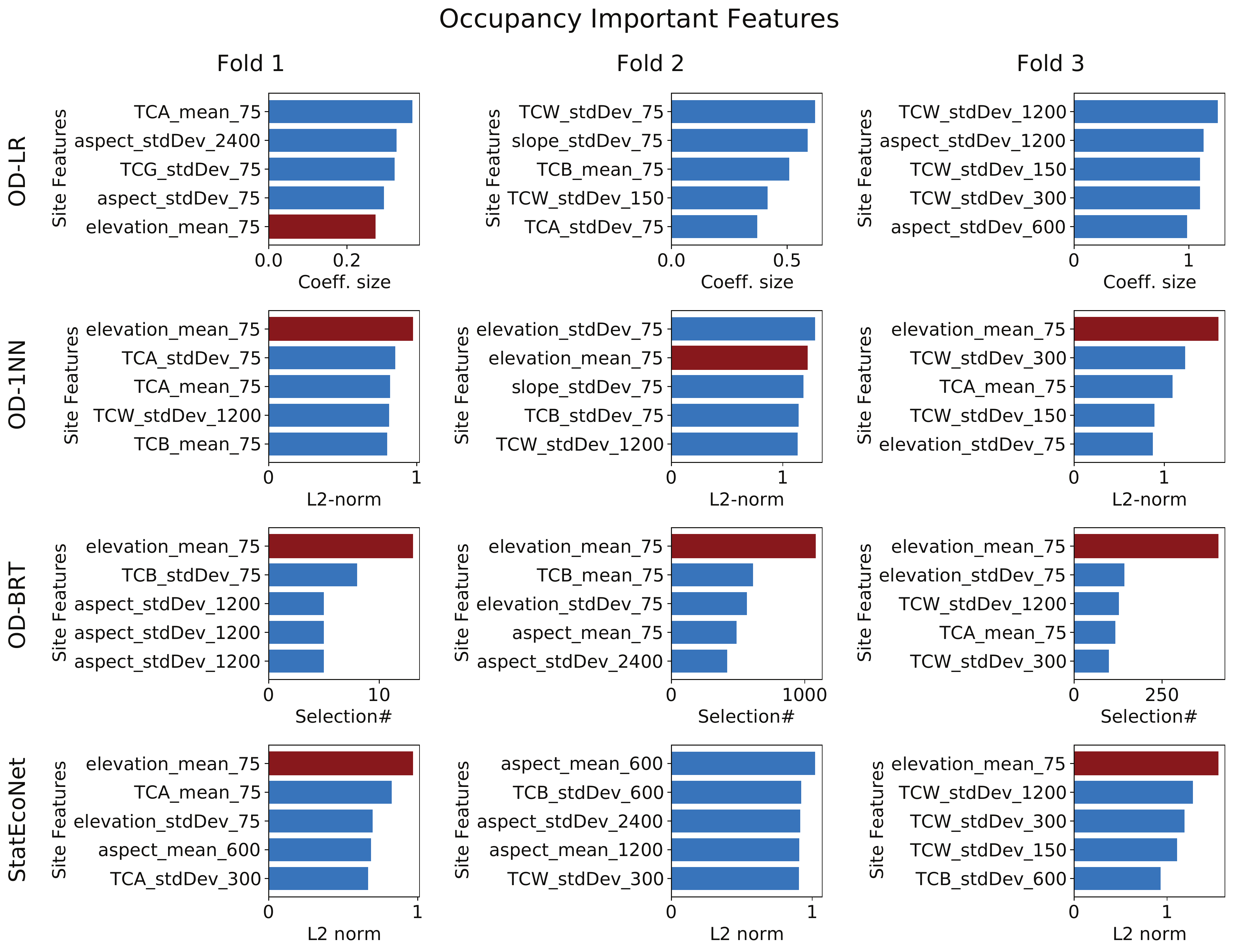}
\caption{Occupancy feature importances for Common Yellowthroat. The top five features per method per fold are plotted. Note that the x-axes differ across methods. The feature corresponding to the mean elevation at the 75 m scale (chosen as an example feature that is important for \texttt{StatEcoNet}) is shaded red to highlight differences across methods.}
\label{fig:COYE_occ_feature}
\end{figure}

\begin{figure}[H]
\centering
  \includegraphics[width=\linewidth]{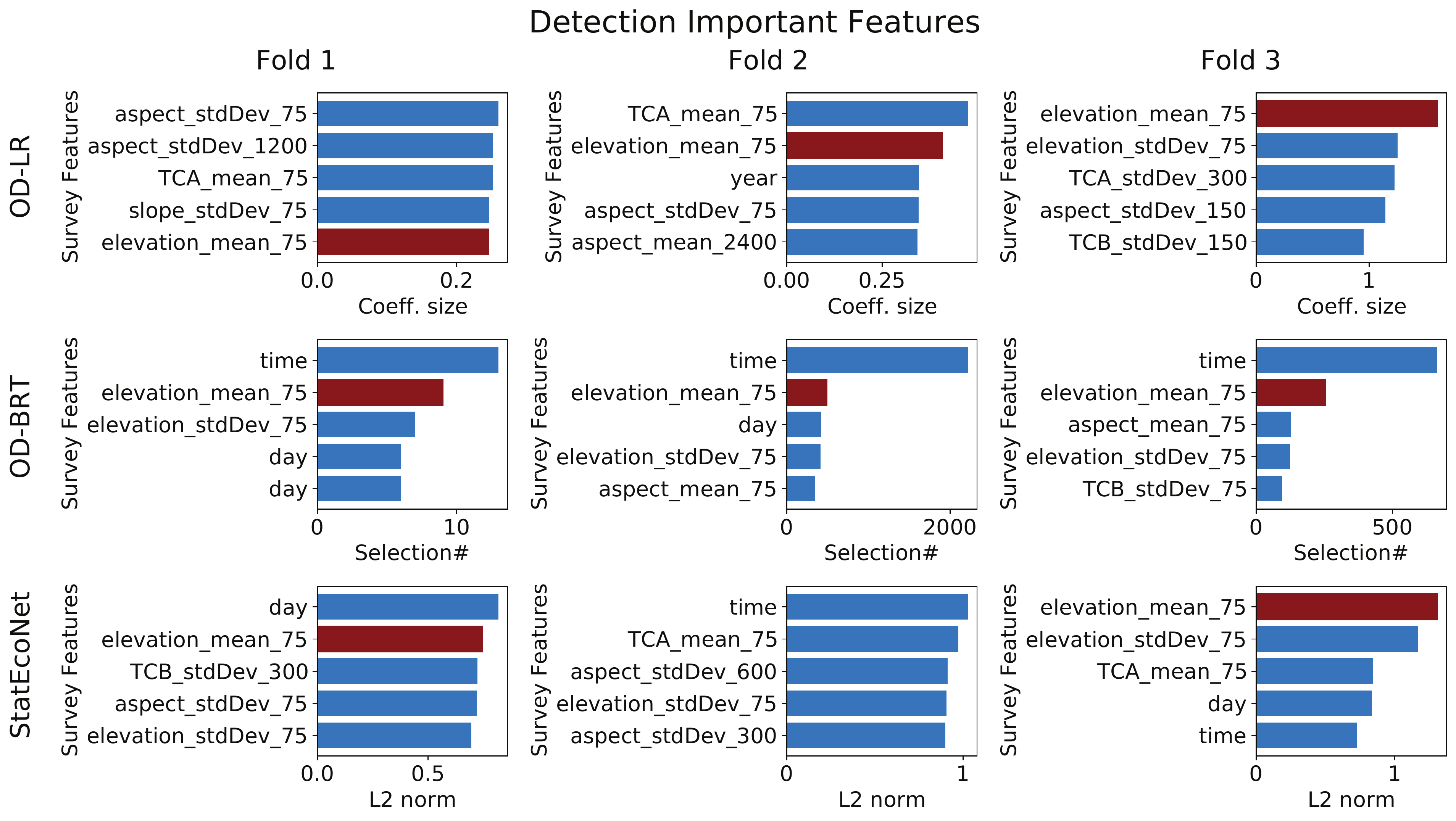}
\caption{Detection feature importances for Common Yellowthroat. The top five features per method per fold are plotted. Note that the x-axes differ across methods. The feature corresponding to the mean elevation at the 75 m scale (chosen as an example feature that is important for \texttt{StatEcoNet}) is shaded red to highlight differences across methods. \texttt{OD-1NN} is not included here because the importance of environmental features to the detection model is not available from that method.}
\label{fig:COYE_det_feature}
\end{figure}

Figures~\ref{fig:COYE_occ_feature} and \ref{fig:COYE_det_feature} show the top five most important variables learned by each method for COYE. Mean elevation was consistently among the top site features, which fits with field observations that this species utilizes wetland and riparian habitats. Such habitats of sufficient size for this species are often found at lower elevations. Inclusion of standard deviations of TCA and TCW probably relate to the contrast between reflectance of water versus adjacent wetland habitats.

\begin{table}[H]
    \centering
    \begin{tabular}{ |c|c|c|c|c| } 
        \hline
        \multirow{2}{*}{Model} & \multirow{2}{*}{Hyper-parameter} & \multicolumn{3}{c|}{Optimal Values} \\
        \cline{3-5}
         &  & Fold 1 & Fold 2 & Fold 3\\
        \hline
        \texttt{OD-LR} & $learningRate$ & 0.01 & 0.01 & 0.01 \\
        \hline
        \texttt{OD-1NN} & $learningRate$ & 0.001 & 0.001 & 0.001 \\
          & $batchSize$ & 32 & 32 & 32 \\
          & $nNeurons$ & 32 & 64 & 16\\
        \hline
        \texttt{StatEcoNet} & $learningRate$ & 0.001 & 0.001 & 0.001\\
          & $batchSize$ & 32 & 32 & 32\\
          & $nNeurons$ & 32 & 64 & 32\\
          & $nLayers$ & 1 & 3 & 1\\
          & $\lambda$ & 0 & 0.001 & 0.001\\
        \hline
         \texttt{OD-BRT} & $shrinkage$ & 0.7274  & 0.4756 & 0.2038\\
         & $bagFraction$ & 0.4633 & 0.9526 &  0.9429\\
          & $treeDepth$ & 3 & 10 & 10\\
        \hline
    \end{tabular}
    \caption{Optimal parameters per fold for Common Yellowthroat}
    \label{tab:opt_parameters_COYE}
\end{table}

\clearpage


\subsection{Eurasian Collared-Dove (EUCD)}
Eurasian Collared-Dove (EUCD) is found in human-dominated habitats. When present, it is usually easy to identify both visually and aurally. However, in noisy urban areas, its calls may be drowned out by other sounds.

\begin{figure}[H]
\centering
  \includegraphics[width=\textwidth]{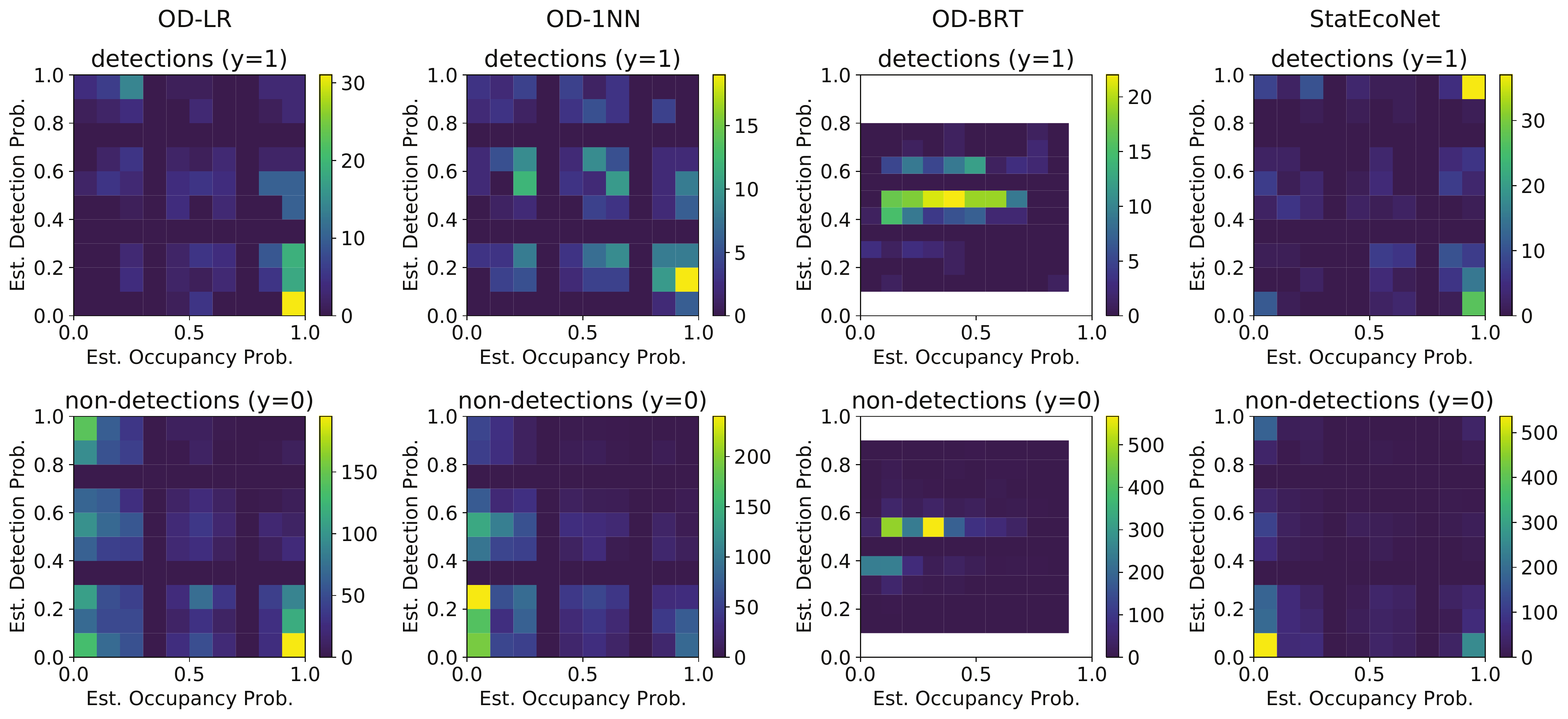}
  \caption{Histograms for Eurasian Collared-Dove.}
  \label{fig:EUCD_densities}
\end{figure}

Figure \ref{fig:EUCD_densities} shows two-dimensional histograms of the occupancy and detection probabilities for all positive species reports (detections, $y=1$) in the top row, and all negative species reports (non-detections, $y=0$) in the bottom row. Here, the bimodality of the occupancy probabilities (\texttt{OD-LR}, \texttt{OD-1NN}, \texttt{StatEcoNet}) makes sense, as human-dominated habitats are relatively easy to distinguish. \texttt{StatEcoNet} shows most detections as having high occupancy and detection probabilities, most non-detections with low occupancy and detection probabilities; this makes sense for a highly-detectable bird with an easily distinguishable habitat. The secondary concentration of sites that are highly likely to be occupied but with very low detection probabilities could be sites where noise pollution impedes detection.

\clearpage 

\begin{figure}[H]
\centering
  \includegraphics[width=\textwidth]{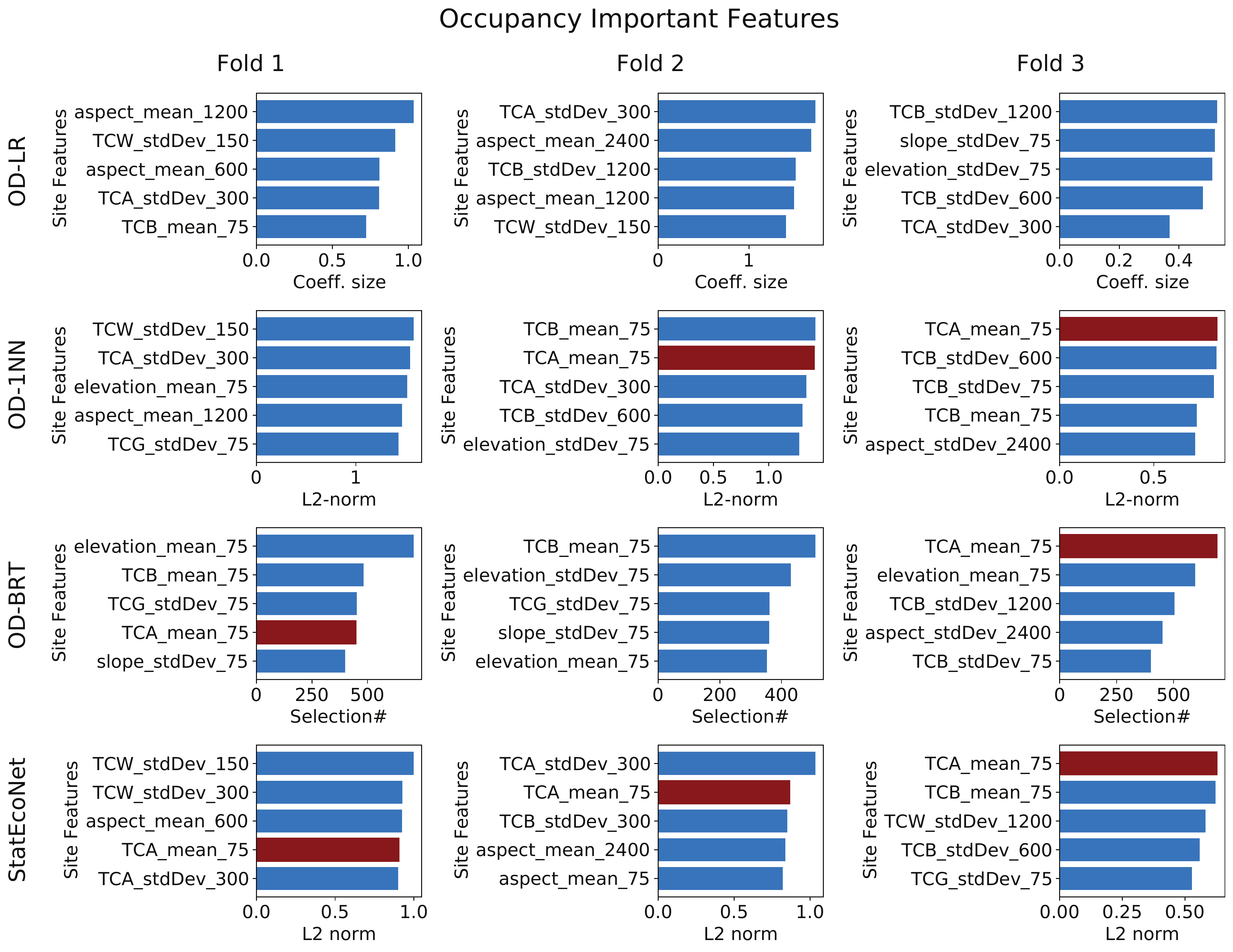}
\caption{Occupancy feature importances for Eurasian Collared-Dove. The top five features per method per fold are plotted. Note that the x-axes differ across methods. The feature corresponding to the mean TCA at the 75 m scale (chosen as an example feature that is important for \texttt{StatEcoNet}) is shaded red to highlight differences across methods.}
\label{fig:EUCD_occ_feature}
\end{figure}

\begin{figure}[H]
\centering
  \includegraphics[width=\textwidth]{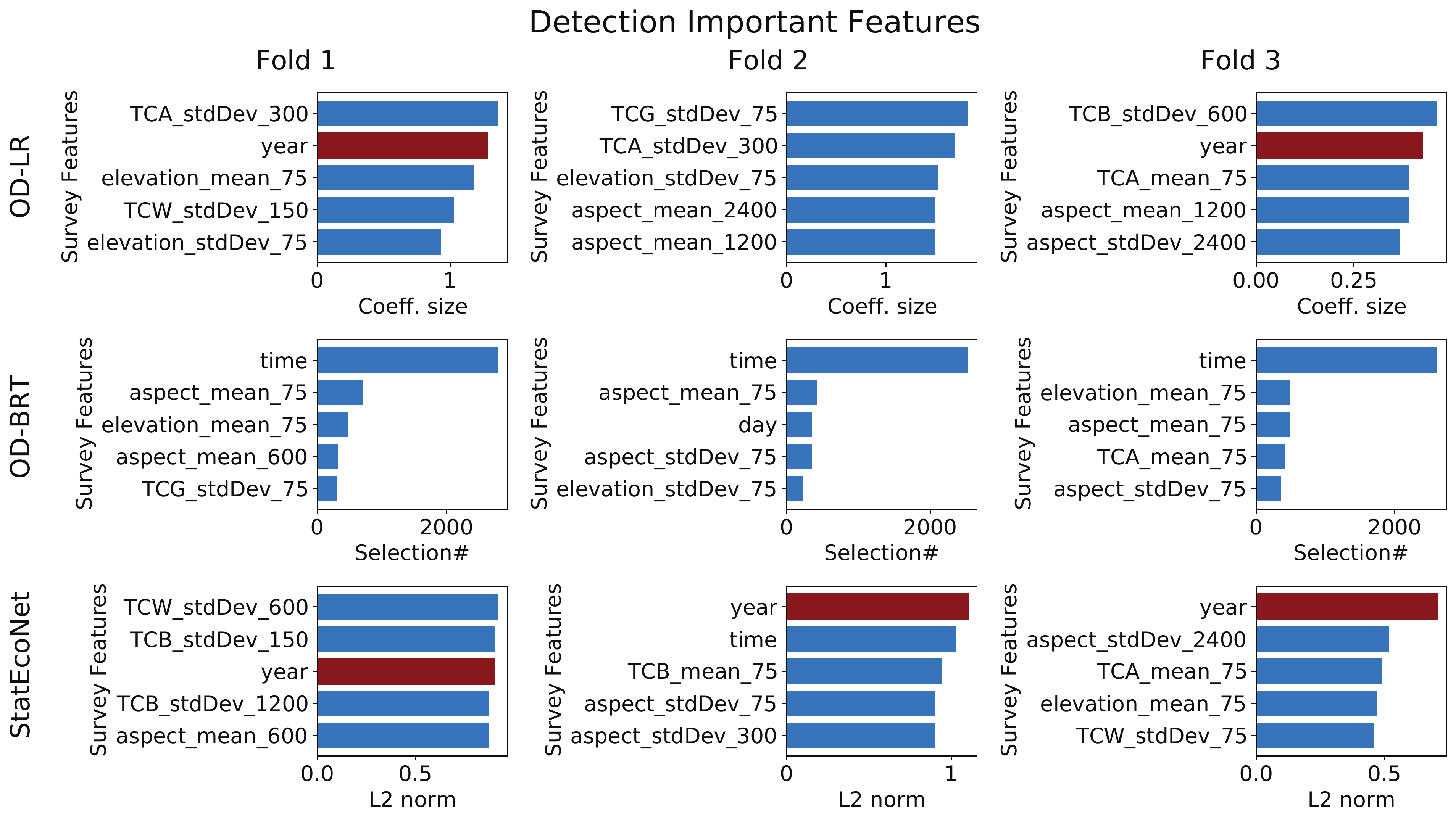}
\caption{Detection feature importances for Eurasian Collared-Dove. The top five features per method per fold are plotted. Note that the x-axes differ across methods. The feature corresponding to the year (chosen as an example feature that is important for \texttt{StatEcoNet}) is shaded red to highlight differences across methods. \texttt{OD-1NN} is not included here because the importance of environmental features to the detection model is not available from that method.}
\label{fig:EUCD_det_feature}
\end{figure}

Figures~\ref{fig:EUCD_occ_feature} and \ref{fig:EUCD_det_feature} show the top five most important variables learned by each method for EUCD. Eurasian Collared-Doves tend to be most numerous around small homesteads (barns, homes) surrounded by agricultural habitats, which is reflected in the identification of TCW standard deviations as important site features. They also are numerous in suburbanized settings, which are captured well by TCA and TCB.

\begin{table}[H]
    \centering
    \begin{tabular}{ |c|c|c|c|c| } 
        \hline
        \multirow{2}{*}{Model} & \multirow{2}{*}{Hyper-parameter} & \multicolumn{3}{c|}{Optimal Values} \\
        \cline{3-5}
         &  & Fold 1 & Fold 2 & Fold 3\\
        \hline
        \texttt{OD-LR} & $learningRate$ & 0.01 & 0.01 & 0.01 \\
        \hline
        \texttt{OD-1NN} & $learningRate$ & 0.001 & 0.001 & 0.001 \\
          & $batchSize$ & $all$ & 32 & 32 \\
          & $nNeurons$ & 64 & 64 & 16\\
        \hline
        \texttt{StatEcoNet} & $learningRate$ & 0.001 & 0.001 & 0.001\\
          & $batchSize$ & 32 & 32 & 32\\
          & $nNeurons$ & 64 & 32 & 16\\
          & $nLayers$ & 3 & 3 & 3\\
          & $\lambda$ & 0.001 & 0 & 0.001\\
        \hline
         \texttt{OD-BRT} & $shrinkage$ & 0.8073  & 0.4250 & 0.8064\\
         & $bagFraction$ & 0.7758 & 0.7055 &  0.9508\\
          & $treeDepth$ & 10 & 7 & 8\\
        \hline
    \end{tabular}
    \caption{Optimal parameters per fold for Eurasian Collared-Dove}
    \label{tab:opt_parameters_EUCD}
\end{table}

\clearpage

\subsection{Song Sparrow (SOSP)}
Song Sparrow (SOSP) is found in most habitats with rich ground-level vegetation. It is usually in wet areas, occasionally restricted to riparian zones, but also found in residential areas with lush vegetation. It can be quite prevalent in some habitats.

\begin{figure}[H]
\centering
  \includegraphics[width=\textwidth]{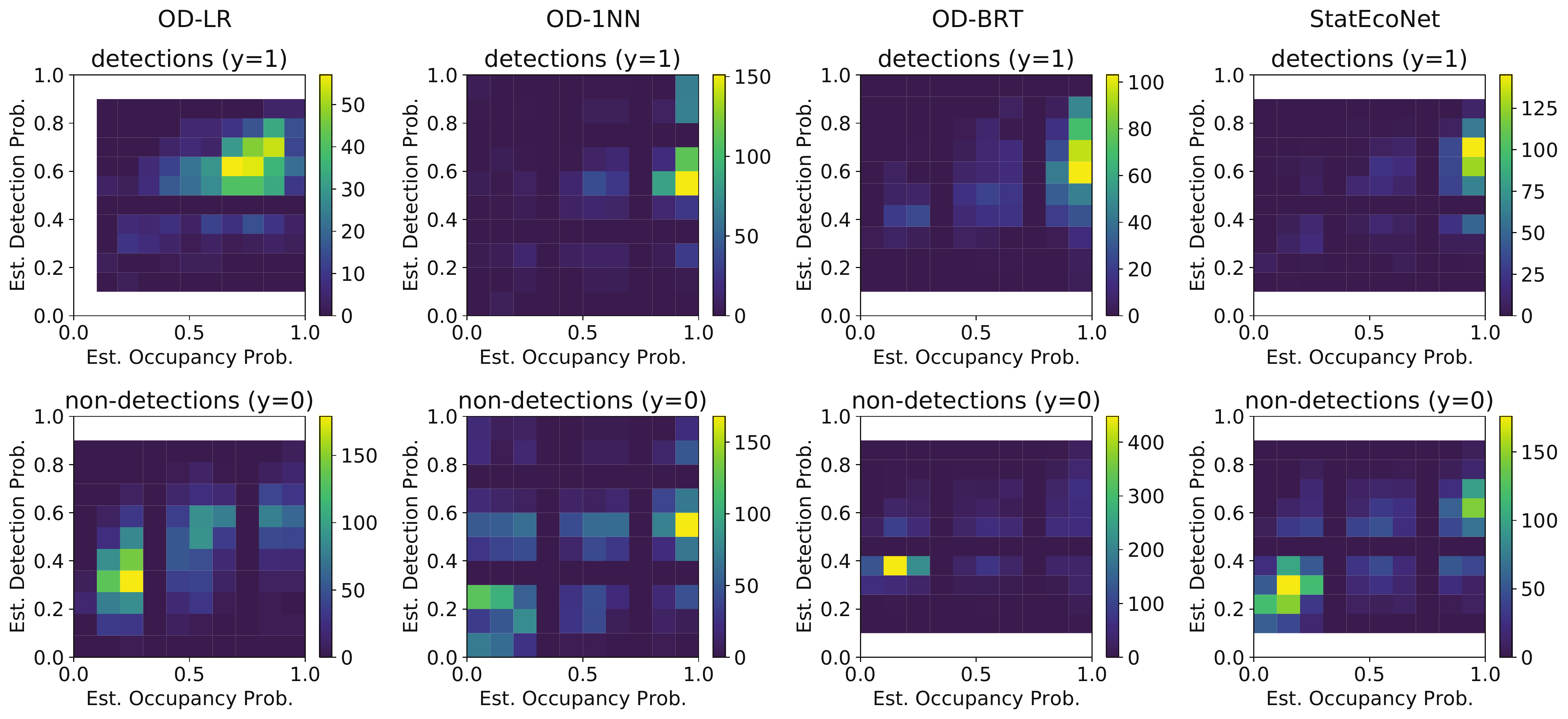}
  \caption{Histograms for Song Sparrow.}
  \label{fig:SOSP_densities}
\end{figure}

Figure \ref{fig:SOSP_densities} shows two-dimensional histograms of the occupancy and detection probabilities for all positive species reports (detections, $y=1$) in the top row, and all negative species reports (non-detections, $y=0$) in the bottom row. Here, the \texttt{OD-BRT} histogram of non-detections concentrating on very low occupancy probabilities seems to imply that almost all of the occupied sites had detections; this is improbable. For the other models, the bimodality of the non-detection occupancy probabilities indicates that the models are finding good separation between habitat and non-habitat and explaining non-detections in good habitat with low detection probabilities.

\clearpage 

\begin{figure}[H]
\centering
  \includegraphics[width=\textwidth]{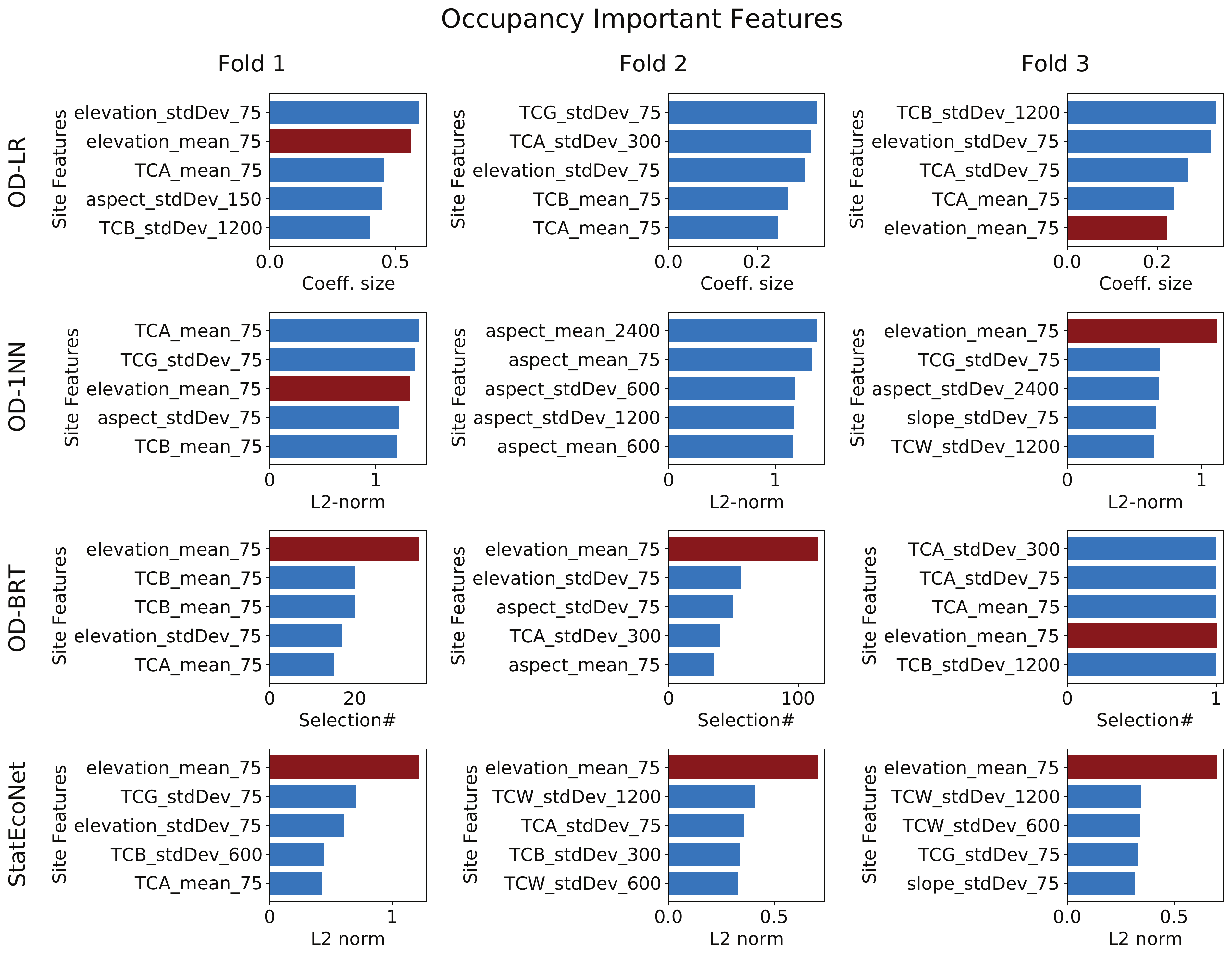}
\caption{Occupancy feature importances for Song Sparrow. The top five features per method per fold are plotted. Note that the x-axes differ across methods. The feature corresponding to the mean elevation at the 75 m scale (chosen as an example feature that is important for \texttt{StatEcoNet}) is shaded red to highlight differences across methods.}
\label{fig:SOSP_occ_feature}
\end{figure}

\begin{figure}[H]
\centering
  \includegraphics[width=\textwidth]{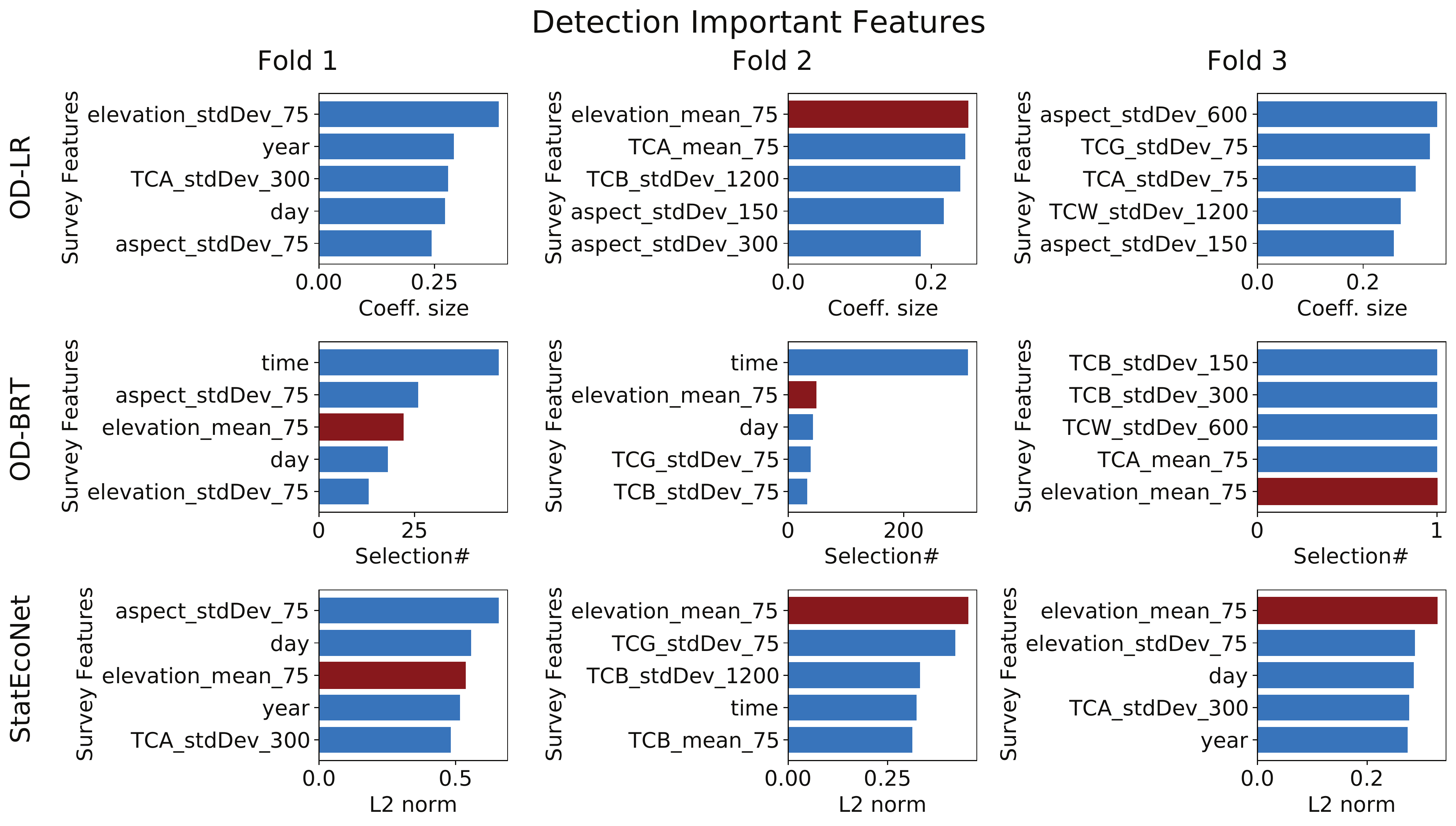}
\caption{Detection feature importances for Song Sparrow. The top five features per method per fold are plotted. Note that the x-axes differ across methods. The feature corresponding to the mean elevation (chosen as an example feature that is important for \texttt{StatEcoNet}) is shaded red to highlight differences across methods. \texttt{OD-1NN} is not included here because the importance of environmental features to the detection model is not available from that method.}
\label{fig:SOSP_det_feature}
\end{figure}

Figures~\ref{fig:SOSP_occ_feature} and \ref{fig:SOSP_det_feature} show the top five most important variables learned by each method for SOSP. Song Sparrows are widely distributed common species associated with riparian habitats, suburban habitats and early successional habitats. Most approaches accurately detected that Song Sparrows most often occur at lower elevations. Because they occupy a wide variety of habitats, specific habitat reflectance features did not consistently emerge across the four analytical approaches, although consistency across the 3 folds was better for \texttt{StatEcoNet}.

\begin{table}[H]
    \centering
    \begin{tabular}{ |c|c|c|c|c| } 
        \hline
        \multirow{2}{*}{Model} & \multirow{2}{*}{Hyper-parameter} & \multicolumn{3}{c|}{Optimal Values} \\
        \cline{3-5}
         &  & Fold 1 & Fold 2 & Fold 3\\
        \hline
        \texttt{OD-LR} & $learningRate$ & 0.01 & 0.01 & 0.01 \\
        \hline
        \texttt{OD-1NN} & $learningRate$ & 0.001 & 0.001 & 0.001 \\
          & $batchSize$ & 32 & $all$ & 32 \\
          & $nNeurons$ & 16 & 32 & 16\\
        \hline
        \texttt{StatEcoNet} & $learningRate$ & 0.001 & 0.001 & 0.001\\
          & $batchSize$ & 32 & 32 & 32\\
          & $nNeurons$ & 64 & 16 & 16\\
          & $nLayers$ & 1 & 3 & 3\\
          & $\lambda$ & 0.01 & 0.01 & 0.01\\
        \hline
         \texttt{OD-BRT} & $shrinkage$ & 0.6779  & 0.6801 & 0.8664\\
         & $bagFraction$ & 0.9158 & 0.2259 &  0.7236\\
          & $treeDepth$ & 3 & 4 & 5\\
        \hline
    \end{tabular}
    \caption{Optimal parameters per fold for  Song Sparrow}
    \label{tab:opt_parameters_SOSP}
\end{table}

\clearpage

\subsection{Western Meadowlark (WEME)}
Western Meadowlark (WEME) strongly specializes on grasslands. Grassland habitat should be more easily distinguishable from our remotely sensed features than some other habitat types (e.g., different types of forest). WEME is one of the most available species for detection in the early morning and can be heard from 1 km away. Since all counts in this dataset were conducted in the morning, high detection probabilities for positive observations make sense.

\begin{figure}[H]
\centering
  \includegraphics[width=\textwidth]{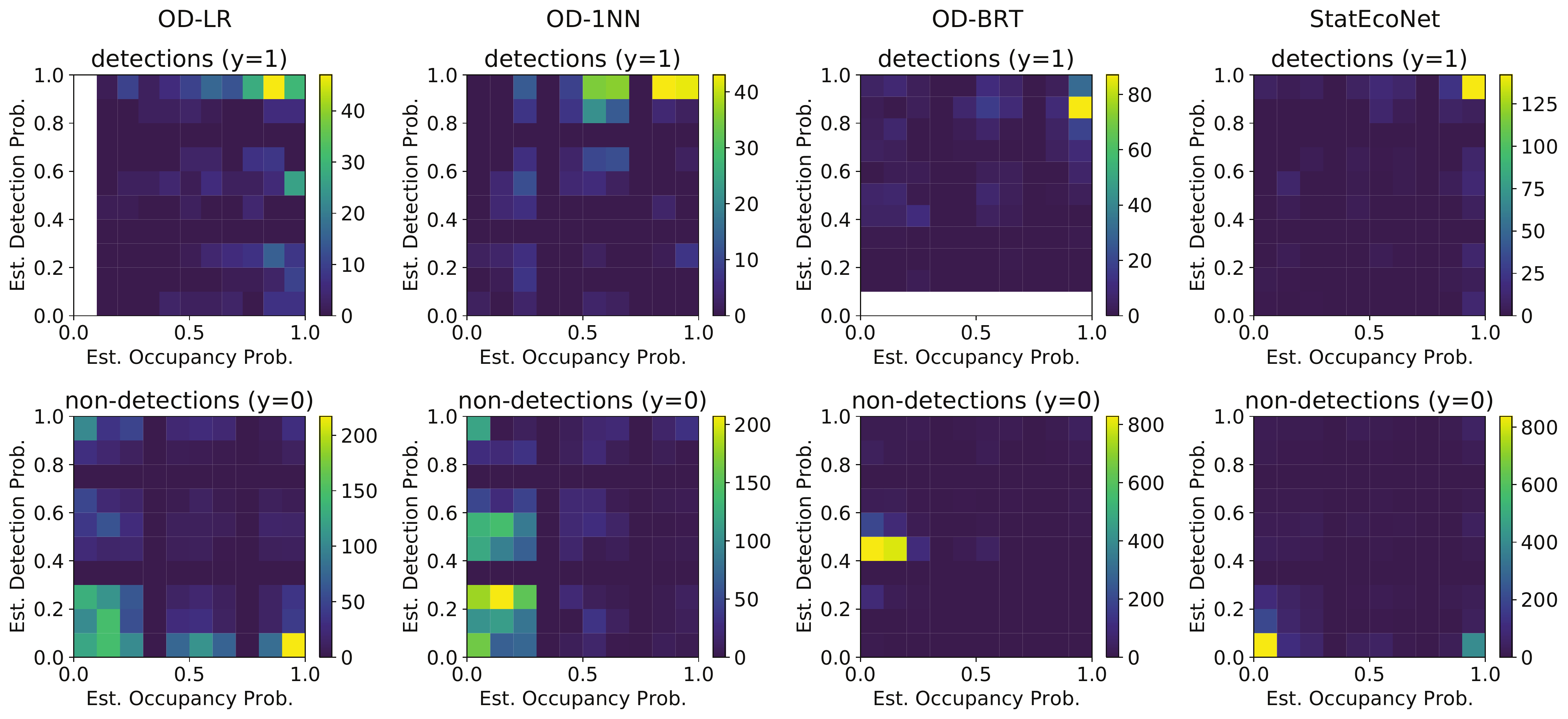}
  \caption{Histograms for Western Meadowlark.}
  \label{fig:WEME_densities}
\end{figure}

Figure \ref{fig:WEME_densities} shows two-dimensional histograms of the occupancy and detection probabilities for all positive species reports (detections, $y=1$) in the top row, and all negative species reports (non-detections, $y=0$) in the bottom row. The \texttt{StatEcoNet} histograms here are quite concentrated, but this may reflect the high detectability of this species and the ease with which its habitat is distinguished by the remote sensing features. The non-detections with high occupancy probability and low detection probability (lower right corner) may be areas of the Willamette valley that do have grassland habitat, but that do not host WEME because they are only small patches of grassland; these would appear to the model as highly likely to be occupied but with low detection probability.

\clearpage 

\begin{figure}[H]
\centering
  \includegraphics[width=\textwidth]{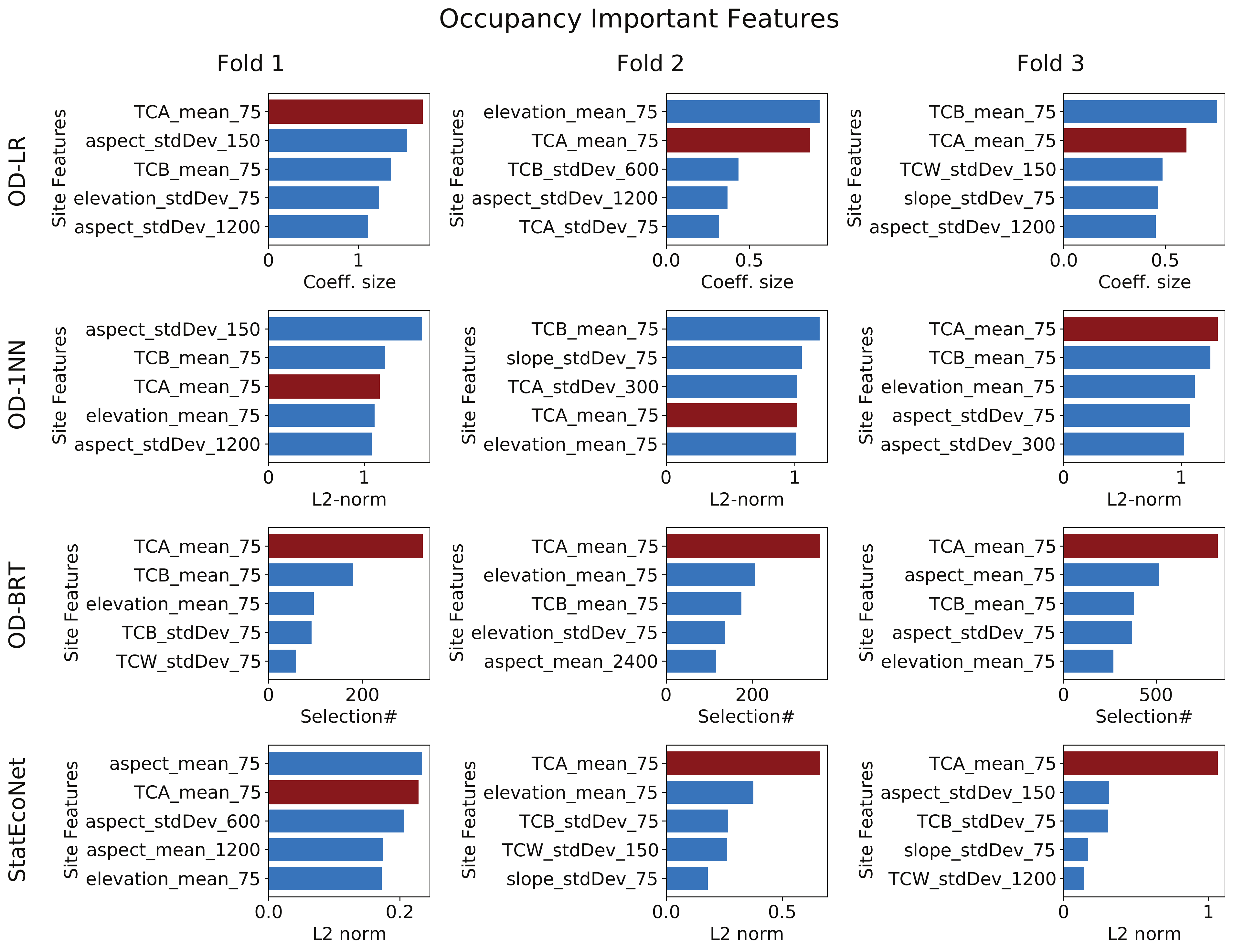}
\caption{Occupancy feature importances for Western Meadowlark. The top five features per method per fold are plotted. Note that the x-axes differ across methods. The feature corresponding to the mean TCA at the 75 m scale (chosen as an example feature that is important for \texttt{StatEcoNet}) is shaded red to highlight differences across methods.}
\label{fig:WEME_occ_feature}
\end{figure}

\begin{figure}[H]
\centering
  \includegraphics[width=\textwidth]{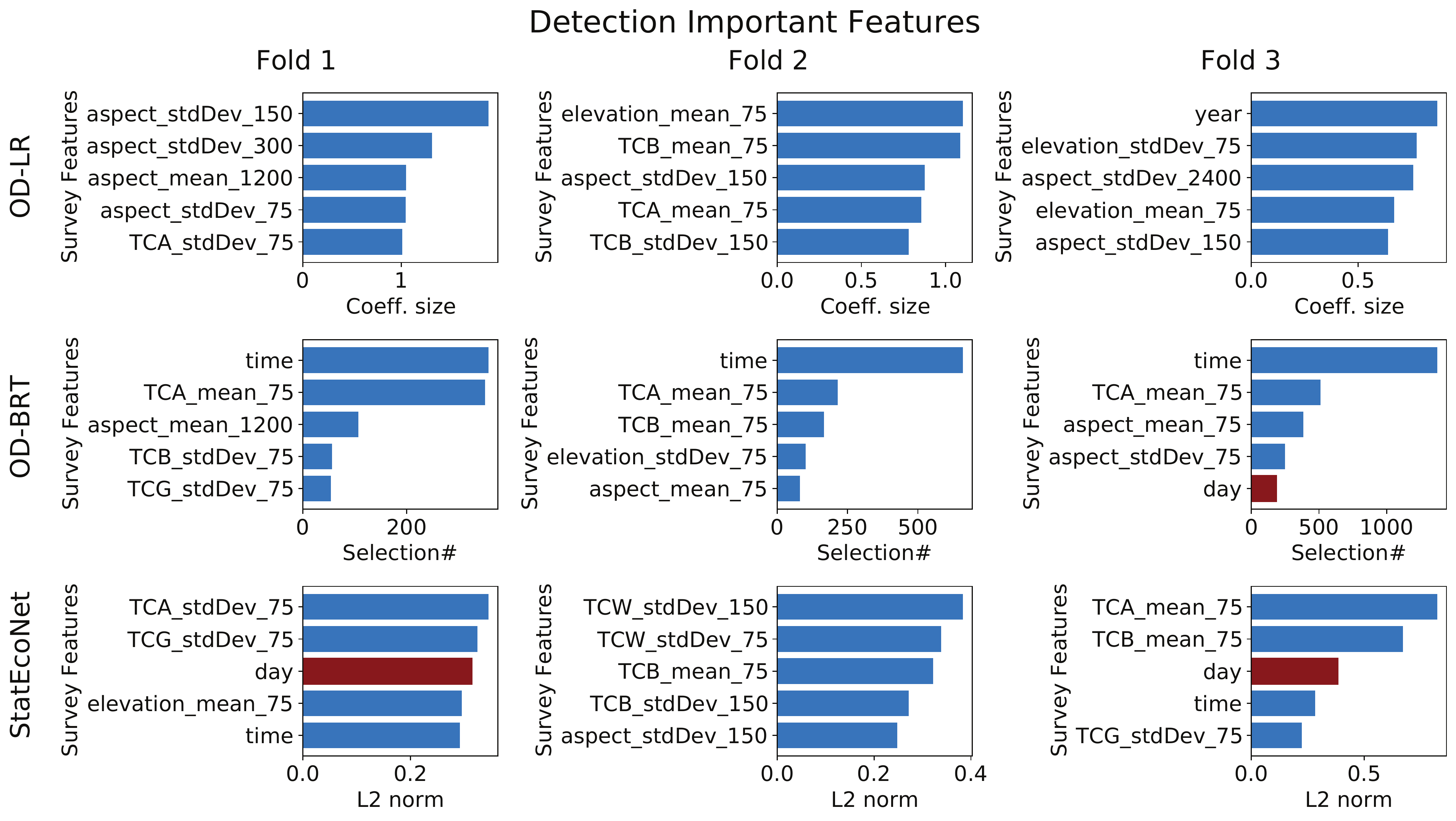}
\caption{Detection feature importances for Western Meadowlark. The top five features per method per fold are plotted. Note that the x-axes differ across methods. The feature corresponding to the day (chosen as an example feature that is important for \texttt{StatEcoNet}) is shaded red to highlight differences across methods. \texttt{OD-1NN} is not included here because the importance of environmental features to the detection model is not available from that method.}
\label{fig:WEME_det_feature}
\end{figure}

Figures~\ref{fig:WEME_occ_feature} and \ref{fig:WEME_det_feature} show the top five most important variables learned by each method for WEME. Western Meadowlarks inhabit grasslands and sagebrush of large extent, avoiding smaller patches or tracts composed largely of agricultural grasslands. Emergence of mean TCA at small buffers (75 m) possibly is related to habitat quality as greenness (moist, productive grasslands) is an important contributor to TCA. 

\begin{table}[H]
    \centering
    \begin{tabular}{ |c|c|c|c|c| } 
        \hline
        \multirow{2}{*}{Model} & \multirow{2}{*}{Hyper-parameter} & \multicolumn{3}{c|}{Optimal Values} \\
        \cline{3-5}
         &  & Fold 1 & Fold 2 & Fold 3\\
        \hline
        \texttt{OD-LR} & $learningRate$ & 0.01 & 0.01 & 0.01 \\
        \hline
        \texttt{OD-1NN} & $learningRate$ & 0.001 & 0.001 & 0.001 \\
          & $batchSize$ & $all$ & 32 & 32 \\
          & $nNeurons$ & 16 & 64 & 64\\
        \hline
        \texttt{StatEcoNet} & $learningRate$ & 0.001 & 0.001 & 0.001\\
          & $batchSize$ & 32 & 32 & 32\\
          & $nNeurons$ & 32 & 64 & 32\\
          & $nLayers$ & 3 & 1 & 1\\
          & $\lambda$ & 0.01 & 0.01 & 0.01\\
        \hline
         \texttt{OD-BRT} & $shrinkage$ & 0.2121  & 0.7199 & 0.4600\\
         & $bagFraction$ & 0.8853 & 0.7763 &  0.2401\\
          & $treeDepth$ & 2 & 6 & 10\\
        \hline
    \end{tabular}
    \caption{Optimal parameters per fold for Western Meadowlark}
    \label{tab:opt_parameters_WEME}
\end{table}

\clearpage

\subsection{Pacific Wren (PAWR)}
\begin{figure}[H]
\centering
  \includegraphics[width=\textwidth]{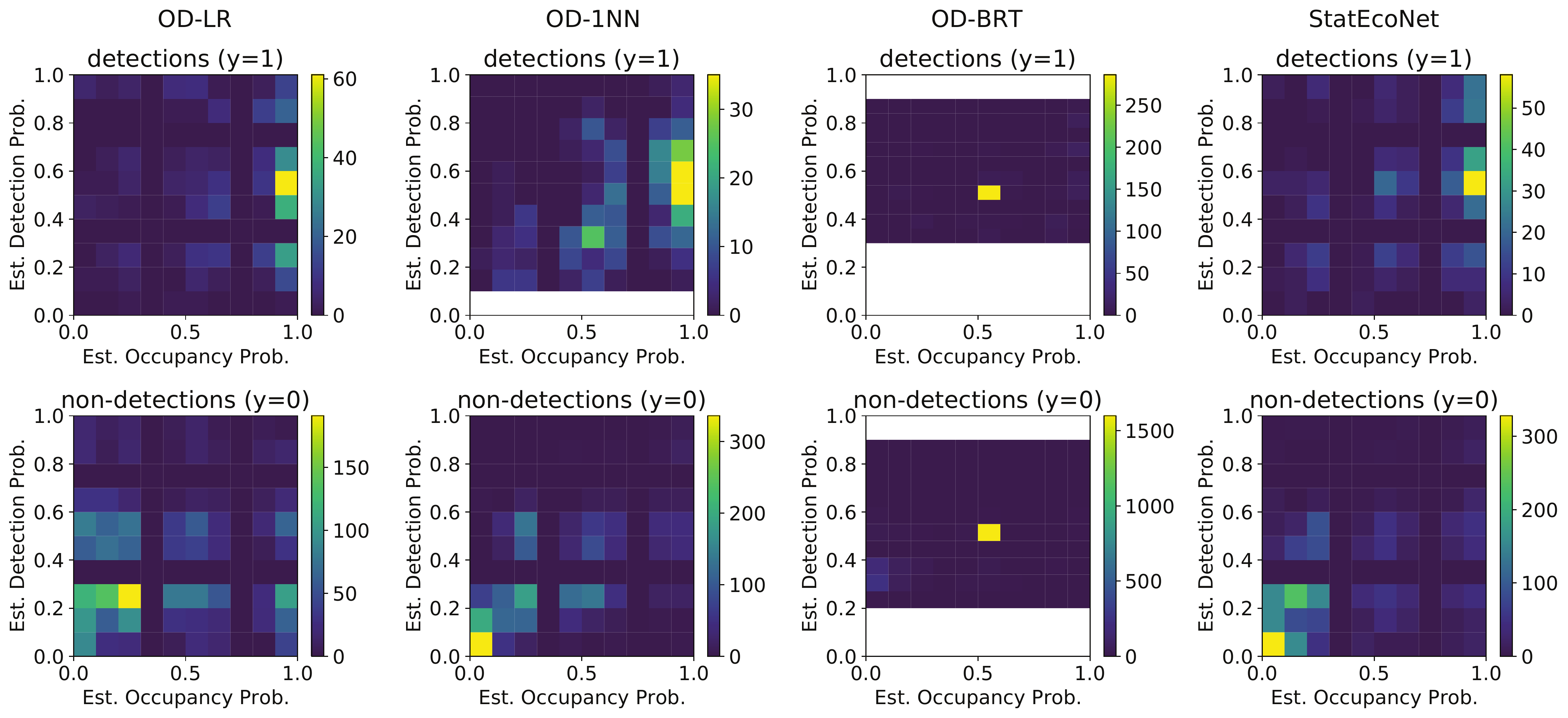}
  \caption{Histograms for Pacific Wren.}
  \label{fig:PAWR_densities}
\end{figure}

This is the example from the main text, repeated here for completeness. In Fig.~\ref{fig:PAWR_densities},
the \texttt{OD-BRT} plots show that many of the model probabilities are highly clustered around 0.5.
This seems to indicate underfitting and is biologically unrealistic. 
The \texttt{OD-LR} and \texttt{OD-1NN} histograms did exhibit high frequencies at the upper right and lower left corners for the detection and non-detection events, respectively. However, the events and the learned models are concentrated in a relatively small number of grid cells, making the histograms spiky. 
This may be pathological since it ties the detected/undetected events with a small number of $\widehat{o}_{i}$ and $\widehat{d}_{it}$---but different sites and surveys may admit a large variety of $\widehat{o}_{i}$ and $\widehat{d}_{it}$ in reality. Hence, although these models could have good estimates for the product $\widehat{o}_i\widehat{d}_{it}$ (and thus similar AUPRCs to \texttt{StatEcoNet}), the individual estimates $\widehat{o}_{i}$ and $\widehat{d}_{it}$ may not be insightful for ecologists.
Encouragingly, the histograms from \texttt{StatEcoNet} show more variability---the probabilities concentrate in the desired regions but also gracefully spread out. This is more likely to be the case in practice.

\clearpage 

\begin{figure}[H]
\centering
  \includegraphics[width=\textwidth]{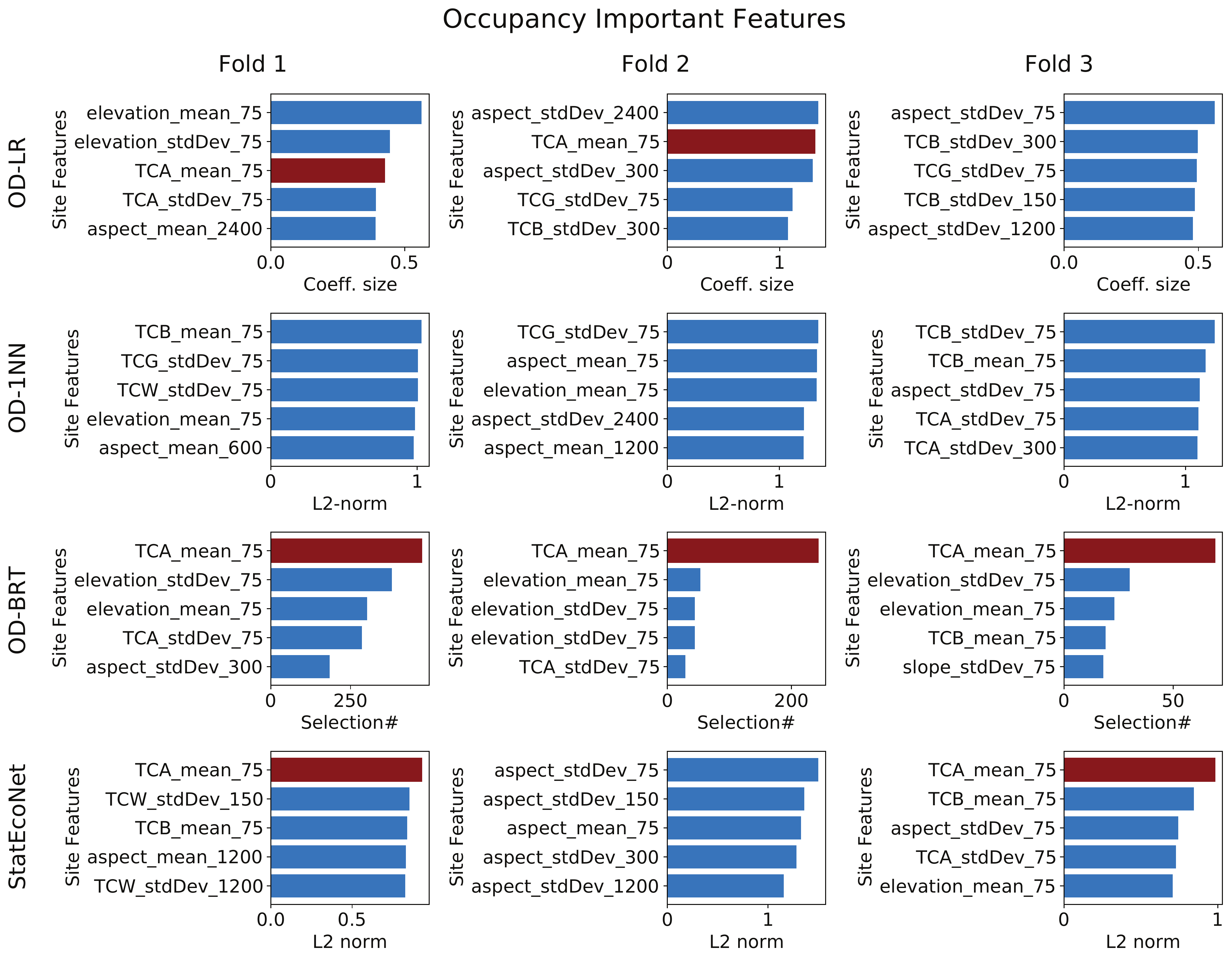}
\caption{Occupancy feature importances for Pacific Wren. The top five features per method per fold are plotted. Note that the x-axes differ across methods. The feature corresponding to the mean TCA at the 75 m scale (chosen as an example feature that is important for \texttt{StatEcoNet}) is shaded red to highlight differences across methods.}
\label{fig:PAWR_occ_feature}
\end{figure}

\begin{figure}[H]
\centering
  \includegraphics[width=\textwidth]{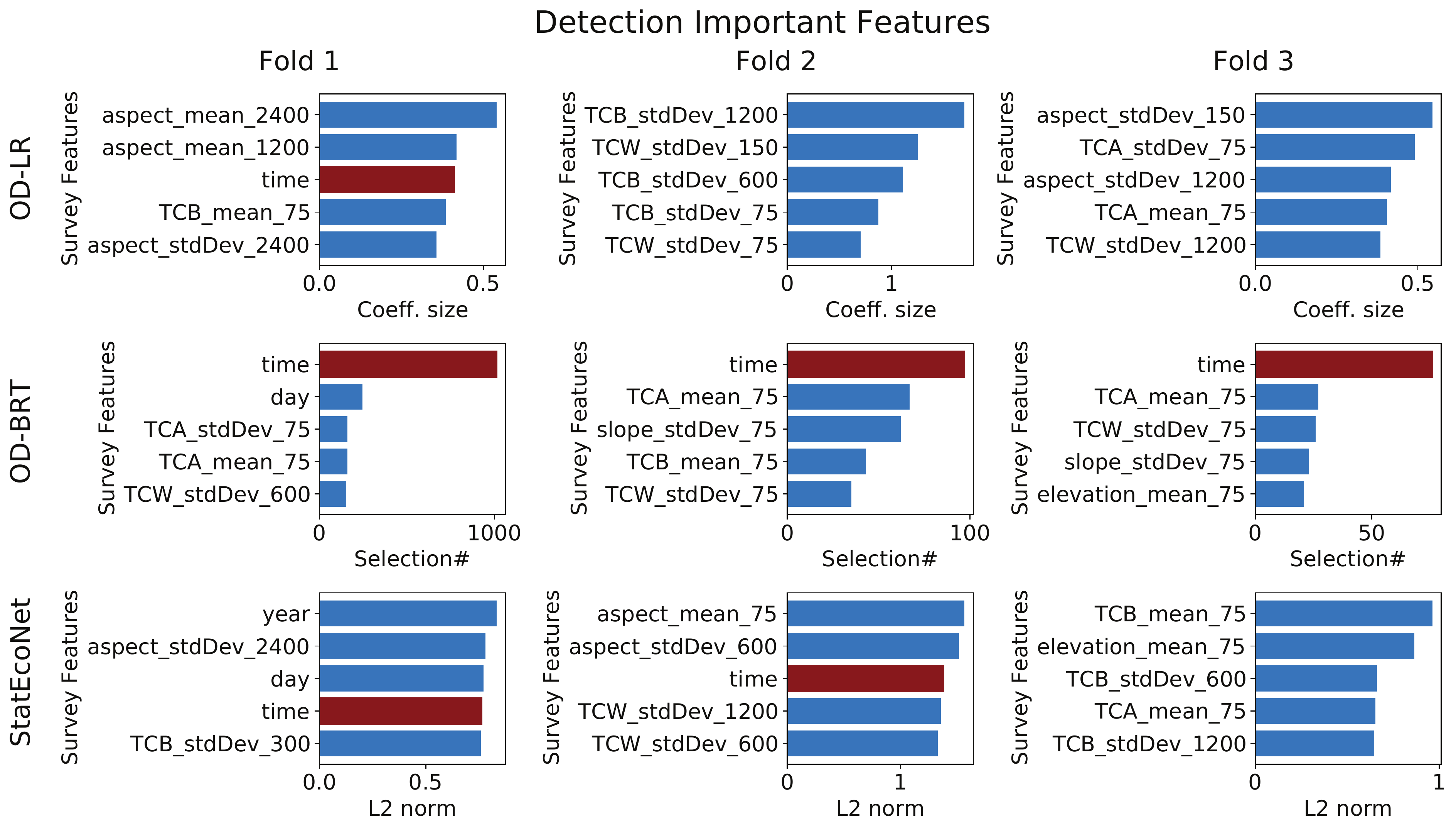}
\caption{Detection feature importances for Pacific Wren. The top five features per method per fold are plotted. Note that the x-axes differ across methods. The feature corresponding to the time (chosen as an example feature that is important for \texttt{StatEcoNet}) is shaded red to highlight differences across methods. \texttt{OD-1NN} is not included here because the importance of environmental features to the detection model is not available from that method.}
\label{fig:PAWR_det_feature}
\end{figure}

Figures~\ref{fig:PAWR_occ_feature} and \ref{fig:PAWR_det_feature} show the top five most important variables learned by each method for PAWR. Inhabiting moist forests, often near riparian zones, Pacific Wrens occupy north-facing slopes that retain moisture later into the dry Pacific Northwest summers. The inclusion of TCA, which captures greenness and brightness, and TCW, capturing correlates of moisture, fits well. The occurrence of aspect also suggests non-random selection of locations in mountainous landscapes by Pacific Wrens.

\begin{table}[H]
    \centering
    \begin{tabular}{ |c|c|c|c|c| } 
        \hline
        \multirow{2}{*}{Model} & \multirow{2}{*}{Hyper-parameter} & \multicolumn{3}{c|}{Optimal Values} \\
        \cline{3-5}
         &  & Fold 1 & Fold 2 & Fold 3\\
        \hline
        \texttt{OD-LR} & $learningRate$ & 0.01 & 0.01 & 0.01 \\
        \hline
        \texttt{OD-1NN} & $learningRate$ & 0.001 & 0.001 & 0.001 \\
          & $batchSize$ &  $all$ & $all$ & $all$ \\
          & $nNeurons$ & 64 & 64 & 32\\
        \hline
        \texttt{StatEcoNet} & $learningRate$ & 0.001 & 0.001 & 0.001\\
          & $batchSize$ & 32 & $all$ & 32\\
          & $nNeurons$ & 64 & 16 & 16\\
          & $nLayers$ & 3 & 1 & 1\\
          & $\lambda$ & 0.01 & 0 & 0.001\\
        \hline
         \texttt{OD-BRT} & $shrinkage$ & 0.100  & 0.100 & 0.4628\\
         & $bagFraction$ & 1.0000 & 0.4268 & 0.3946\\
          & $treeDepth$ & 4 & 2 & 3\\
        \hline
    \end{tabular}
    \caption{Optimal parameters per fold for Pacific Wren}
    \label{tab:opt_parameters_PAWR}
\end{table}

\clearpage

\section{Computing Infrastructure}
\begin{table}[H]
    \centering
    \begin{tabular}{ |c|l|l|l| } 
        \hline        
        \multirow{5}{*}{Hardware} & \multirow{3}{*}{CPU} & \# of Cores & 4\\
                \cline{3-4}
                 &     & \# of Threads & 8 \\
                 \cline{3-4}
                 &     & Model & Intel(R) Xeon(R) CPU E3-1230 v5 @ 3.40GHz \\
                 \cline{2-4}
                 & Memory & \multicolumn{2}{l|}{16 GB} \\
                 \cline{2-4}
                 & Operating System & \multicolumn{2}{l|}{CentOS Linux 7} \\
        \hline
        \hline
        \hline
        \multirow{24}{*}{Software}  & \multicolumn{2}{l|}{Python} & 3.8.3 \\
                  \cline{2-4}
                  & \multirow{9}{*}{Python libraries} & torch & 1.5.1 \\
                  & & numpy & 1.19.1  \\
                  & & pandas & 1.0.5  \\
                  & & matplotlib & 3.3.0 \\
                  & & tqdm & 4.48.0  \\
                  & & scikit-learn & 0.23.1  \\
                  & & scipy & 1.5.2  \\
                  & & jupyterlab & 2.2.1  \\
                  & & import-ipynb & 0.1.3  \\
                  \cline{2-4}
                  & \multicolumn{2}{l|}{R} & 4.0.2 \\
                  \cline{2-4}
                  & \multirow{13}{*}{R libraries} & grt & 0.2.1 \\
                  &  & reshape2 & 1.4.4 \\
                  &  & PRROC & 1.3.1 \\
                  &  & Metrics & 0.1.4 \\
                  &  & paramtest & 0.1.0 \\
                  &  & Rcpp & 1.0.5 \\
                  &  & scales & 1.1.1 \\
                  &  & dplyr & 1.0.1 \\
                  &  & ggplot2 & 3.3.2\\
                  &  & patchwork & 1.0.1 \\
                  &  & blockCV & 2.1.1 \\
                  &  & raster & 3.3-13 \\
                  &  & sf & 0.9-5\\
                  
        \hline
    \end{tabular}
    \caption{Computing infrastructure specification.}
    \label{tab:computing_infrastructure}
\end{table}

\clearpage
\bibliography{supplement.bib}
\bibliographystyle{aaai21}